\documentclass{article}
    \usepackage[nonatbib,final]{neurips_2024}

\usepackage[utf8]{inputenc} % allow utf-8 input
\usepackage[T1]{fontenc}    % use 8-bit T1 fonts
\usepackage{hyperref}       % hyperlinks
\usepackage{url}            % simple URL typesetting
\usepackage{booktabs}       % professional-quality tables
\usepackage{amsfonts}       % blackboard math symbols
\usepackage{nicefrac}       % compact symbols for 1/2, etc.
\usepackage{microtype}      % microtypography
\usepackage{xcolor}         % colors
\usepackage{subfiles}
\usepackage{graphicx}
\usepackage{amsmath}
\usepackage{multirow}

\newcommand{\xhdr}[1]{\vspace{2mm}\noindent{{\bf #1.}}}
\newcommand{\modelname}{Maia-2}

\title{Maia-2: A Unified Model for Human-AI Alignment in Chess}

\author{
  Zhenwei Tang \\
  University of Toronto \\
  \texttt{josephtang@cs.toronto.edu} \\
  \And
  Difan Jiao \\
  University of Toronto \\
  \texttt{difanjiao@cs.toronto.edu} \\
  \And
  Reid McIlroy-Young \\
  Harvard University \\
  \texttt{reidmcy@seas.harvard.edu} \\
  \And
  Jon Kleinberg \\
  Cornell University \\
  \texttt{kleinberg@cornell.edu} \\
  \And
  Siddhartha Sen \\
  Microsoft Research \\
  \texttt{sidsen@microsoft.com} \\
  \And
  Ashton Anderson \\
  University of Toronto \\
  \texttt{ashton@cs.toronto.edu} \\
}

\begin{document}

\maketitle

\begin{abstract}

    % Artificial intelligence has demonstrated superhuman performance across various domains. Chess, with its rich history as a pivotal testbed for AI research, offers a unique perspective on human--AI alignment, i.e., AI for mimicking human chess moves instead of outperforming human players, allowing us to enhance chess learning experiences with more relatable AI partners and deepen our insights into human decision-making.
    % Efforts have been made towards chess AI for matching human moves, while they are fragmented models for different strength levels with isolated predictions, which lead to a lack of coherence in the models' ability to adapt and respond across the full spectrum of player skills, ultimately limiting their effectiveness as learning tools and AI partners.
    % In this work, we propose a unified framework for human--AI alignment in chess that reacts coherently across a wide range of strength levels.
    % Recognizing the complex, non-linear nature of players' learning and application of chess strategies and tactics as they progress, we introduce a Strength-Aware Attention mechanism to dynamically integrate players' strengths with encoded chess positions, enabling our model to be reactive to the evolving strengths of players.
    % Extensive experimental results demonstrate that our unified framework significantly enhances the alignment between AI and human players across a diverse range of chess expertise, paving the way for more relative AI partners and insights into human decision-making.

There are an increasing number of domains in which artificial intelligence (AI) systems both surpass human ability and accurately model human behavior. This introduces the possibility of algorithmically-informed teaching in these domains through more relatable AI partners and deeper insights into human decision-making. Critical to achieving this goal, however, is coherently modeling human behavior at various skill levels. Chess is an ideal model system for conducting research into this kind of human-AI alignment, with its rich history as a pivotal testbed for AI research, mature superhuman AI systems like AlphaZero, and precise measurements of skill via chess rating systems. Previous work in modeling human decision-making in chess uses completely independent models to capture human style at different skill levels, meaning they lack coherence in their ability to adapt to the full spectrum of human improvement and are ultimately limited in their effectiveness as AI partners and teaching tools. In this work, we propose a unified modeling approach for human-AI alignment in chess that coherently captures human style across different skill levels and directly captures how people improve. Recognizing the complex, non-linear nature of human learning, we introduce a \emph{skill-aware attention mechanism} to dynamically integrate players’ strengths with encoded chess positions, enabling our model to be sensitive to evolving player skill. Our experimental results demonstrate that this unified framework significantly enhances the alignment between AI and human players across a diverse range of expertise levels, paving the way for deeper insights into human decision-making and AI-guided teaching tools. 
Our implementation is available \href{https://github.com/CSSLab/maia2}{here}.
    \end{abstract}

\section{Introduction}

%Want AI partners and learning tools.
There are an increasing number of domains in which artificial intelligence (AI) systems both surpass human ability and accurately model human behavior. 
This combination of machine mastery over a domain and computational understanding of human behavior in it introduces the possibility of algorithmically-informed teaching and learning. 
AI-powered aids could guide people along reliable and efficient improvement paths, synthesized from their knowledge of both human trajectories and objective performance. 
Relatable AI partners, on the other hand, could learn to act alongside human counterparts in synergistic and complementary ways.

% Chess is a great domain.
Researchers have begun to tackle this challenge in the model system of chess. 
Once held to be an ideal testbed for developing artificial intelligence, it is now the perfect domain to pursue human-AI alignment. 
The AI community finally surpassed all human ability in chess approximately 20 years ago, a milestone achievement and watershed cultural moment. 
Now, superhuman AI chess engines are ubiquitous and widely used. 
Despite this transformation, chess has never been more popular, becoming a mainstream activity in many countries during the last few years. 
There is now both unprecedented demand for chess education, as well as mature superhuman AI that could in principle help meet it. 

% But existing models aren't good learning tools because they aren't coherent. 
However, % despite the existence of superhuman AI in chess, 
existing models fall short of being effective learning tools and relatable partners. 
Traditional chess engines such as Stockfish and AlphaZero are unimaginably strong, but they don't play in ways that humans can easily understand or learn from. 
Comparing one's own decisions with those of traditional engines, it is easy to see how near-perfect AI would have improved upon your play but hard to see how you could realistically do the same. 
Recent work has resulted in the development of Maia, a suite of models that aim to mimic human behavior in chess at various skill levels by learning to predict actual human moves from a wealth of online gameplay data~\cite{mcilroy2020aligning}. 
While substantially more human-like, these models still cannot power effective algorithmic teaching tools because of several limitations.

% Want unified, coherent, accurate model.
First and foremost, Maia models players at different skill levels completely independently; games by players at one skill level and those by an adjacent skill level are fed into separate instances of the same architecture and result in separate models. 
This has the downside that predictions from one model are independently made of predictions from any other. 
Viewed as a whole, they are volatile: the Maia models might predict that at one level players will approach a position correctly, then at the next level they will make a horrible mistake, then at the next level they will do fine again, and so on.
In a word, they fail to \emph{cohere}. 
People don't improve along volatile paths, they steadily get better. 
The unrealistically incoherent predictions made by separate models don't suggest realistic pathways that people can take in order to get better. 
In order to serve as algorithmic teachers or learning aids, our models of human behavior must be coherent.

% Hard because chess players are insanely variable in their skill, chess is deep, capturing one of the widest stretches of ability in a single model is a challenge. 
% Important because coherence is how people actually are. Want to ask counterfactual questions like what do I have to change in order to get better. 
Building a coherent model of human skill in chess is difficult, because the breadth of skill in chess is almost incomprehensibly large. 
Decisions made by beginners bear only the faintest of relations to those made by masters. 
A difference of 200 points in chess rating systems roughly equates to a 75\% win rate for the higher-rated player---typically higher than the best record of any team in the entire National Basketball Association. 
On the online chess platform we study, there are players who are 2600 rating points apart---or 13 successive steps of 75\%-vs.-25\% dominance apart from each other. 
Capturing this breadth of skill in a single model, in a coherent, smooth fashion, is a challenge.

% What we do: skill-aware attention.
We contribute a unified modeling approach for human-AI alignment in chess that coherently captures human style across different skill levels and directly captures how people improve. 
Since our model builds directly on original Maia, we call it \emph{Maia-2}. 
Maia-2 consists of a standard residual network tower that processes chess positions into features, and our novel contribution of a \emph{skill-aware attention module} with channel-wise patching. 
This innovation takes the position representation outputted by the residual network tower and simple player skill encodings and learns how player skill levels interact with chess positions to produce the moves humans make. 
Unlike previous models, Maia-2 only requires the current board position as input (as opposed to six), which dramatically reduces training time and increases flexibility (e.g.\ for applying the model in non-game contexts where there may be no 6-board history). 
In addition to policy and value heads like in previous work, we also add an additional auxiliary information head that helps the model learn a deeper understanding of human chess moves.

% Our results: accuracy and coherence. 
We evaluate Maia-2 along two key dimensions: move prediction \emph{accuracy} and \emph{coherence}. 
Testing it against the original Maia models, Stockfish, and AlphaZero, Maia-2 emerges as the most accurate human move predictor by far, surpassing original Maia by almost 2 full percentage points. 
Analyzing move prediction accuracy by skill level, Maia-2 matches and surpasses all other models on all skill levels. 
Furthermore, Maia-2's gains in perplexity are similarly striking, reducing average perplexity from a previous record of 4.67 bits down to 4.07 bits. 
Maia-2 achieves these accuracy gains while being substantially more coherent than the original Maia models. 
For example, call a model's treatment of a position \emph{monotonic} if it assigns a monotonically increasing probability to the correct move as we increase skill. 
While original Maia treats 1\% of a random sample of positions monotonically, Maia-2 treats a remarkable 27\% of the same positions monotonically. 
This is in keeping with our intuitive understanding of how chess players steadily and smoothly improve across the skill range. 
Finally, we conduct an investigation of the human chess concepts Maia-2 learns and varies with skill via linear probes, and find that skill-dependent concepts like overall board evaluation indeed vary with skill, but skill-independent concepts do not, which also accords with our understanding of how human players make decisions.

%\reid{Intro bullet points}

% \begin{itemize}
%     \item AI systems that act like humans are becoming common and highly sought after, additionally AI systems are in many cases super-human (Games, image recognition, trip planning, ...)
%     \item Beyond using them for rough replacements we also want them as collaborators, teachers, etc
%     \item How can we leverage these AI systems to become useful teachers 
%     \item As a first step we need AIs that can understand the breadth of human skill, we need \textit{skill-aware} AI
%     \item Chess is great model system for this thanks to the long history in AI research, and lots of well labeled data
%     \item Previous work has created models that can do move prediction, but they are limited to specific skill ranges and cannot model variations in skill
%     \item There is also work being done to explore how these models work (probes), which we engage with by allowing another dimension to the analysis
%     \item In this paper we present a higher performing model that allows for more control over the skill and is designed to work as a base for future progress
%     \item We also have have higher accuracy and a cool result with the probes
% \end{itemize}
\section{Related Work}

\xhdr{Chess and AI} This paper draws on the long history of chess at the forefront of AI research~\cite{hsu1999ibm,clark1999sciences,turochamp,shannon2001mathematical}. We engage with 3 distinct approaches to building chess AI: heuristic~\cite{stockfish}, learned~\cite{silver2016mastering}, and textual~\cite{feng2023chessgpt}. %These methods all aim for achieving high levels of performance in the game. 
\textit{Heuristic search:} The original approach to computer chess was heuristics-based%for evaluating the game state then search for the best actions that result in the best future position, à la the Minimax theorem
~\cite{turochamp,hsu2002behind}. This method was famously used by IBM's Deep Blue to defeat Garry Kasparov~\cite{hsu1999ibm} and is currently used by Stockfish~\cite{stockfish}, one of the strongest chess engines in the world. %The use of heuristics makes these methods much more interpretable than the next type.
\textit{Learned search:} Alpha(Zero) Go~\cite{silver2016mastering,silver2017mastering} is a set of neural networks that learn to play Go with methods that generalized to other games, including chess, with AlphaZero~\cite{silver2017mastering}. 
Chess AI with learned search is also extended to multi-agent systems~\cite{zahavy2023diversifying}, where diverse AI systems can outperform a single AI in challenging tasks such as chess.
\textit{Chess as text:} Large language models~\cite{vaswani2017attention,brown2020language,touvron2023llama} have recently been found to perform well on tasks that the models were not \textit{explicitly} trained on~\cite{kojima2022large}, including playing chess without fine-tuning~\cite{wu2023autogen,noauthor_llmchess_nodate,carlini_carlinichess-llm_2024}. This has lead to chess knowledge being one of the tested features in BIG-Bench~\cite{srivastava2022beyond}, a popular LLM evaluation suite. Additionally, fine-tuning a language model can lead to systems that not only play chess, but can also generate comments, describe positions, and create other simple analyses of a game~\cite{lee2022improving,noever2020chess,feng2023chessgpt}. 
%The more recent method~\cite{feng2023chessgpt} makes significant improvements, by using a larger dataset and more parameters, but struggle to perform reliably with even simple tasks such as correctly predicting annotations the model barely exceeds random guessing.

\xhdr{Human-AI Alignment in Chess} Building a chess engine that can defeat any human has been a solved problem for over 20 years. 
%now almost trivial thanks to the hard work of generations of computer scientists describe above. 
This has led to a new research agenda in extracting useful knowledge from these superhuman systems. 
A direct way of doing this is to probe an AI chess engine in a human representation space. 
Without any prior human knowledge or guidance, evidence of human chess concepts learned by AlphaZero is found and measured by linear probes~\cite{mcgrath2022acquisition}. Going further, AlphaZero also encodes knowledge that extends beyond existing human knowledge but is ultimately learnable by humans~\cite{schut2023bridging}.%, from a cooperative experiment between human chess grandmasters and AlphaZero . 
Another direction was the creation of a `behavioral stylometry' model that can identify chess players from the moves they play~\cite{mcilroy2021detecting}. %embedding chess players based on a small (100) number of games. The method has some correlation with skill, but the resulting embedding is not easily used for tasks beyond direct identification.
%These studies thereby bridge the gap between superhuman AI strategies and human-understandable chess knowledge.
An alternative approach to creating systems that can act as guides to humans is demonstrated by Maia~\cite{mcilroy2020aligning,mcilroy2022learning}, in which a model is trained to predict the next move a human will play, instead of optimizing for winning the game.
In addition to predicting human actions the models have been fine-tuned to predict a given player's actions~\cite{mcilroy2022learning}. 
% This method suggests that players have a distinct and measurable style that can be learned by observing a sufficient number of games. 
The prediction accuracy can be improved via a reinforcement learning-style search~\cite{jacob2022modeling}. %The regularization factor allows for a trade-off between the base policy suggested by the neural network and the new policy suggested by search, with more emphasise on search performing better at stronger skill levels.

\section{Methodology}
\label{sec:methodology}
%\subsection{Overview}

\begin{figure}[!t]
\vspace{-0.2cm}
	\centering
	\includegraphics[width=0.9\textwidth]{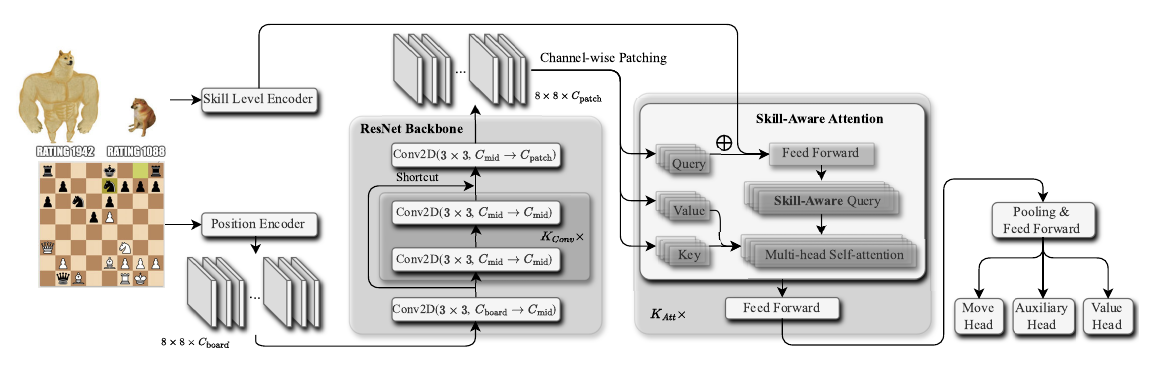}
 \vspace{-0.3cm}
        \caption{Overview of the Maia-2 model architecture.}
  \label{fig:model}
\vspace{-0.5cm}
\end{figure}

We propose a unified model architecture to capture human decision-making in chess across a broad spectrum of skill levels. Since this model builds upon the previous Maia move-matching models, we call it \emph{Maia-2}. 
As shown in Figure~\ref{fig:model}, Maia-2 first encodes active and opponent skill levels and the chess positions, respectively. Then the encoded skill levels and positions are fused using our skill-aware attention with channel-wise patching architecture. The fused representations are then used for move prediction (policy head), auxiliary information prediction (auxiliary head), and game outcome prediction (value head). We now discuss each of these components in detail.

\subsection{Skill Level Encoder}
\label{section:skill_level_encoder}
Instead of directly incorporating player ratings as numerical inputs, we use categorical skill level embeddings for two reasons. First, 
player behavior and decision-making in chess are not linearly related to their rating.
Categorical embeddings allow for capturing complex, non-linear relationships between player strength and their moves. They can encode nuanced differences in play style and strategy that are not directly proportional to player ratings.
Second, Generalization across similar skill levels: Players within a certain skill level may exhibit similar playing styles, strategies, and common mistakes. Categorical embeddings group players into these ranges, helping the model to better generalize across players with similar strengths, as opposed to treating each rating as a numerical input.

Let $\mathbf{E} \in \mathbb{R}^{|\mathbf{E}| \times d_s}$ be the matrix of player rating embeddings, where each row corresponds to the embedding of a skill level with dimension $d_s$: $\mathbf{E} = [\mathbf{e}_{(0, 1000]}, \mathbf{e}_{(1000, 1100]}, ..., \mathbf{e}_{(2000, +\infty)} ]^\top$.
% \begin{equation}
%     \mathbf{E} = [\mathbf{e}_{(0, 1000]}, \mathbf{e}_{(1000, 1100]}, ..., \mathbf{e}_{(2000, +\infty)} ]^\top.
% \end{equation}
Given the skill levels $a$ and $o$ of an \underline{a}ctive player (i.e.\ the player to move) and the \underline{o}pponent player, we look up the embedding matrix $\mathbf{E}$ by rows to map the skill levels to active and opponent skill embeddings: $\mathbf{e}_{a} = \mathbf{E}[a], \mathbf{e}_{o} = \mathbf{E}[o].$
% \begin{equation}
%     \mathbf{e}_{a} = \mathbf{E}[a], \quad \mathbf{e}_{o} = \mathbf{E}[o].
% \end{equation}

Note that previous work~\cite{mcilroy2020aligning, jacob2022modeling} uses completely independent models for human-AI alignment at different skill levels---e.g.\ decisions by 1100-rated chess players are encoded in one model and decisions by 1500-rated players are encoded in a separate model. 
%to targeting at the same active and opponent skill levels.
Further, these models ignore opponent skill level, meaning that predictions cannot vary as a function of opponent strength.
However, the active player's decisions may be significantly affected by the opponent's skill level in certain types of situations, or even in general. 
Players may adjust their strategy based on their perception of the opponent's skill, e.g.\ a higher-skill opponent might prompt more (or less) cautious play, while against a lower-skill opponent a player may pursue more aggressive tactics. 
Thus, the interaction between the skill levels of both players is an important component of matching human moves.
Unlike existing models that ignore opponent skill level (and actually only consider games in which both players are at the same skill level), we explicitly model not only opponent skill but also the complex interplay between the two players' skill levels, and how it affects human decision-making. %model consider the pairs of active and the opponent player's skill levels to model the more nuanced skill-aware move matching.

\subsection{Position Encoder}
\xhdr{Position representation}
We use a well-established method~\cite{silver2017mastering, mcilroy2020aligning} to represent each chess position as a multi-channel tensor $P_{\text{input}} \in \mathbb{R}^{C_{\text{board}}\times8\times8}$, which includes channels for each type of chess piece, which color is to move, and states of the position that are not derivable from the position alone (castling rights and en passant), where $C_{\text{board}}$ denotes the number of channels.
One important departure from previous work is that we only use the current chess position, and not the last few chess positions that occurred in the game (models have typically incorporated the six most recent positions in the game). Many games with perfect information, including chess, can be modeled as alternating Markov games~\cite{littman1994markov, silver2016mastering}, where future states are independent of past states given the current game state.
Therefore, the current chess position theoretically encapsulates all the information necessary to make future decisions. 
Although human decision-making in chess may sometimes subtly depend on the historical lead-up to the current position, these effects are anecdotally small. %This characteristic makes historical positions redundant for determining human moves.
In exchange, we gain two large practical benefits. First, modeling AI-human move matching in a Markovian way vastly improves training \textit{efficiency} by reducing the computational load via significantly smaller data usage for each decision. Second, it also enhances \textit{flexibility}, enabling our resulting model to make predictions even without historical data, which is particularly advantageous in situations where only the current position is available, like chess training puzzles or any position that didn't necessarily occur in a full game.

\xhdr{Position encoding}
To process the position representation $P_{\text{input}}$, we encode $P_{\text{input}}$ with the well-established ResNet-based~\cite{he2016deep} backbone architecture for chess position modeling with $K_{\text{Conv}}$ sequentially connected blocks~\cite{mcilroy2020aligning}: $P_{\text{encoded}} = \text{Backbone}_{\times K_{\text{Conv}}}(P_{\text{input}}) \in \mathbb{R}^{C_{\text{patch}}\times8\times8}$,
% \begin{equation}
%     P_{\text{encoded}} = \text{Backbone}_{\times K_{\text{Conv}}}(P_{\text{input}}) \in \mathbb{R}^{C_{\text{patch}}\times8\times8},
% \end{equation}
where $P_{\text{encoded}}$ denotes the encoded position representation of $C_{\text{patch}}$ channels.
More details about position representation and the backbone architecture can be found in Appendix Section~\ref{appendix:channels}.

\subsection{Bridging Skill Levels and Positions}
\label{section:attention}
% \joseph{TODO: Explain bridging these two is non-trivial, emphasize on the complexity and non linearity of skill levels}

A central challenge we face is learning how players at different skill levels interact with chess positions differently. How does an expert player evaluate and process a chess position to come up with a move, and how does this differ from a novice? The relationship between positions and skill levels is complicated by the non-linearity in how players of various skill levels interpret and react to chess positions. This complexity presents a significant challenge in human move prediction using a unified model for diverse skill levels. To bridge skill levels and positions---decision-makers and decisions---we propose \emph{skill-aware attention with channel-wise patching}.

\xhdr{Channel-wise patching}
% \joseph{TODO: why channel-wise? cite leela trials on ViT}
In contrast to the area-wise patching approach in Vision Transformers (ViTs)~\cite{dosovitskiy2020image}, we employ channel-wise patching. Each channel is flattened and linearly transformed, regarding the number of channels in $P_{\text{encoded}}$, i.e., $C_{\text{patch}}$, as the sequence length: $P_{\text{patched}} = \text{Patching}(P_{\text{encoded}}) \in \mathbb{R}^{C_{\text{patch}}\times64}$, $P = P_{\text{patched}}\mathbf{W} + \mathbf{b} \in \mathbb{R}^{C_{\text{patch}}\times d_{\text{att}}}$,
% \begin{align*}
%     &,\\
%     &,
% \end{align*}
where $\mathbf{W} \in \mathbb{R}^{64\times d_{\text{att}}}$ and $\mathbf{b} \in \mathbb{R}^{d_{\text{att}}}$ denote the parameters of the linear projection from the patching dimension to the hidden dimension of the skill-aware attention blocks $d_{\text{att}}$.
This is particularly suitable for patching encoded chess positions as inputs to Transformer-like architectures, where channels are essentially feature maps that represent different learned latent concepts. These concepts in feature maps are then interactively selected and aggregated considering skill levels via skill-aware attention.

% \xhdr{Channel-wise patching}
% % \joseph{TODO: why channel-wise? cite leela trials on ViT}
% In contrast to the area-wise patching approach in Vision Transformers (ViTs)~\cite{dosovitskiy2020image}, we employ channel-wise patching. Each channel is flattened and linearly transformed, regarding the number of channels in $P_{\text{encoded}}$, i.e., $C_{\text{patch}}$, as the sequence length.
% \begin{align*}
%     &P_{\text{patched}} = \text{Patching}(P_{\text{encoded}}) \in \mathbb{R}^{C_{\text{patch}}\times64},\\
%     &P = P_{\text{patched}}\mathbf{W} + \mathbf{b} \in \mathbb{R}^{C_{\text{patch}}\times d_{\text{att}}},
% \end{align*}
% where $\mathbf{W} \in \mathbb{R}^{64\times d_{\text{att}}}$ and $\mathbf{b} \in \mathbb{R}^{d_{\text{att}}}$ denote the parameters of the linear projection from the patching dimension to the hidden dimension of the skill-aware attention blocks $d_{\text{att}}$.
% This is particularly suitable for patching encoded chess positions as inputs to Transformer-like architectures, where channels are essentially feature maps that represent different learned latent concepts. These concepts in feature maps are then interactively selected and aggregated considering skill levels via skill-aware attention.

\xhdr{Skill-aware Attention}
Given position representations $P_{\text{patched}}$ and skill level representations $\mathbf{e}_{a}$ and $\mathbf{e}_{o}$, our proposed skill-aware multi-head self-attention is computed as follows. For each head \( k \), we learn weight matrices \( \mathbf{W}^Q_k \in \mathbb{R}^{d_{\text{att}} \times d_h} \), \( \mathbf{W}^K_k \in \mathbb{R}^{d_{\text{att}} \times d_h} \), and \( \mathbf{W}^V_k \in \mathbb{R}^{d_{\text{att}} \times d_h} \), where $d_h$ denote the dimension of each head. The queries \( Q_k \), keys \( K_k \), and values \( V_k \) for each head are computed as: $Q_k = P_{\text{patched}}\mathbf{W}^Q_k$, $K_k = P_{\text{patched}}\mathbf{W}^K_k$, $V_k = P_{\text{patched}}\mathbf{W}^V_k$.
% \begin{equation*}
%     Q_k = P_{\text{patched}}\mathbf{W}^Q_k, \quad K_k = P_{\text{patched}}\mathbf{W}^K_k, \quad V_k = P_{\text{patched}}\mathbf{W}^V_k
% \end{equation*}
In order to fuse player skill levels and chess positions progressively and interactively, we inject skill level embeddings into queries within the multi-head self-attention: $Q^*_k = Q_k + (\mathbf{e}_{a} \oplus \mathbf{e}_{o})\mathbf{W}^*,$
% \begin{equation}
%     Q^*_k = Q_k + (\mathbf{e}_{a} \oplus \mathbf{e}_{o})\mathbf{W}^*,
% \end{equation}
where $\mathbf{W}^* \in \mathcal{R}^{2d_s \times d_h}$ denotes the weight matrix for feature transformation to the query space, and $\oplus$ is the concatenation operator.
We choose to incorporate skill levels in queries because queries directly influence how attention is distributed across patched channels. Using skill-aware queries $Q^*_k$, the attention mechanism can adjust its focus to reflect the strategic considerations and positional understanding of players at different skill levels. This adjustment allows \modelname{} to adaptively prioritize features of the positions that are more relevant to the skill levels involved, enhancing the model's contextual sensitivity.
The skill-aware scaled dot-product attention for each head is thus defined as: $h_k = \text{softmax}\left(\frac{Q^*_kK_k^T}{\sqrt{d_k}}\right)V_k$.
% \begin{equation*}
%     h_k = \text{softmax}\left(\frac{Q^*_kK_k^T}{\sqrt{d_k}}\right)V_k
% \end{equation*}
The outputs of all heads $h_1, {h}_2, \ldots, {h}_h$ are concatenated and then linearly transformed: $P_{\text{att}} = \sigma(({h}_1 \oplus {h}_2 \oplus \ldots \oplus {h}_h)\mathbf{W}^O)$,
% \begin{equation}
%     P_{\text{att}} = \sigma(({h}_1 \oplus {h}_2 \oplus \ldots \oplus {h}_h)\mathbf{W}^O),
% \end{equation}
where \( \mathbf{W}^O \in \mathbb{R}^{hd_h \times d_{\text{att}}} \) denote the weight matrix for multi-head attention and $\sigma(\cdot)$ denotes the activation function. We apply the vanilla ViT's feed-forward network and add \& norm components upon $P_{\text{att}}$ to obtain the output of each skill-aware attention block $P_{\text{out}}$.
In \modelname{}, we employ a sequence of skill-aware attention blocks to progressively fuse skill levels and positions. Specifically, the output $P_{\text{out}}$ for the previous block is fed into the next block as the input. We denote the final output after $K_{\text{Att}}$ blocks as $P$. This procedure enables the model to refine its understanding and interpretation of the positions with each successive block.

\subsection{Model Training}

\xhdr{Infusing auxiliary information}
\label{section:aux}
To enhance the model's understanding of the game state, we inject auxiliary information as labels, including \textit{legal moves} represented by multi-hot vectors and \textit{human move information}: one-hot vectors of which piece is moved, which piece is captured (if any), the move's originating square, the move's destination square, and whether or not the move will deliver a check.
% For nuanced moves like castling, we ensure both the king's and rook's moves are accurately represented with 2-hot vectors. 
These segments are used as labels for classification, serving a dual purpose: 1) It offers a more granular understanding of human moves by providing detailed context beyond just the move indices produced by the policy head labels, enriching the model's insight of player decisions; and 2) It ensures the model also learns about objective (i.e.\ chess-specific as opposed to behavioral) knowledge in chess, which is essential for developing a comprehensive understanding of both human moves and the fundamental mechanics of the game.

\xhdr{Data balancing and filtering}
Chess games between players of significantly different skill levels are relatively rare but help us understand how players of lower skill levels  approach games against far stronger opponents and vice versa.
While previous work has ignored these games completely, they play a central role in our approach. 
Since games between players of similar skill levels vastly outnumber more uneven matchups, we use a data balancing strategy to effectively train our unified model for aligning players across all skill levels, in which games between players of different skill levels are over-sampled.
Online chess platforms feature a variety of game types, including blitz, rapid, and classical, each representing games played at different time controls (amount of time given to each player for the whole game). 
We focus on Rapid games, which are medium-length games that lie between the fast-paced decisions of ``Blitz'' games and the slower, more strategic considerations of ``Classical'' games. In addition, we follow the procedures in~\cite{mcilroy2020aligning} to filter valid positions within each game. 
More details about data balancing and filtering are available in Appendix~\ref{appendix:detail}.

\xhdr{Training objectives}
With the fused skill level and position representation $P$ as input, we construct the policy head on top to predict human moves, which is optimized using cross-entropy loss with one-hot labels representing the recorded human move.
We also build the auxiliary information head to infuse additional knowledge into \modelname{} as introduced in Section~\ref{section:aux}. This head is trained using bit-wise binary cross-entropy loss with multi-hot labels. %The policy head is optimized  
Finally, following previous work~\cite{mcilroy2020aligning, mcilroy2022learning} we include a value head to predict the game outcome as a regression task, where the labels 1, 0, -1 denote winning, drawing, and losing, respectively. The training objectives of these heads are balanced to contribute equally to \modelname{} model optimization.
Hyperparameter settings used for \modelname{} training can be found in Appendix Table~\ref{tab:hyperparameter}.

\section{Results}
We empirically evaluate \modelname{} along two key dimensions: move prediction \emph{accuracy}, how well it can predict human moves at varying skill levels, and move prediction \emph{coherence}, how aligned its predictions are across skill levels.
% \xhdr{\modelname{}}
We train \modelname{} 
% with the methodology described in Section~\ref{sec:methodology} 
on Lichess games played between Jan 2013 and Nov 2023, with the exception of December 2019, since that is the month used for testing in the original Maia paper (and we also test on this month for consistency)~\cite{mcilroy2020aligning}. After game filtering and balancing, we end up with a training set of 169M games (9.1B positions).
% played between May 2018 and November 2023 (inclusive).
% To control for differences in training sets when comparing \modelname{} against Maia, w
We also train 
\modelname{}$_{\text{subset}}$ with identical model architecture and training configurations as \modelname{}, except it only has access to the same training data that Maia had for fair comparisons.
% (i.e.\ games played from Jan 2013 to Nov 2019 in the \href{https://database.lichess.org/}{Lichess Database}).
%Since Maia benchmark data was built on the games in Dec 2019, we skipped these data in \modelname{} training for fair comparisons. 
Dataset statistics are reported in Appendix Tables~\ref{tab:trainingdata_stats}, \ref{tab:balanceddata_stats_subset}, and \ref{tab:balanceddata_stats}.
% \xhdr{Model comparison}
We compare \modelname{} with Stockfish~\cite{stockfish}, the strongest chess engine, 
% in the world at the time of writing. Since the most common method of using Stockfish to play in a ``human-like'' way is to limit its search depth, we test it at various search depths. The second is  
\href{https://lczero.org/}{Leela}, an open-source counterpart to AlphaZero~\cite{silver2017mastering}.
% which we also test at various strengths, since chess commentators have anecdotally observed that its decisions are more reminiscent of human-like play at lower strengths.
% Finally, our main model comparison will be with 
and Maia~\cite{mcilroy2020aligning},
% (which we will refer to as ``Maia'' to avoid confusion with our model), 
the state-of-the-art model for human-like chess play. Maia is actually a set of 9 separate models, each trained on a different set of players at different skill levels from 1100 to 1900. 
% Maia-1100 models the weaker players, Maia-1500 the intermediate players, and Maia-1900 the higher-skill players. 
% \textcolor{orange}{Joseph: TODO: explain why 1599, Lichess ratings start at 1500, as is recommended by the Glicko system definition, 2000 is top 10 percent
%As the human move prediction performance comparisons shown in Table~\ref{tab:baseline}, we have the following observations:
% \xhdr{Evaluation Datasets}
\label{section:testset}
% We evaluate using three main test sets. 
% To enhance fair comparison with baseline models, w
We use the benchmarking \textit{Maia Testset} ~\cite{mcilroy2020aligning} for performance comparisons where both players have identical skill levels. 
We report the results on \textit{Maia Testset} by grouping players into three categories: \textit{Skilled} (Rapid rating up to 1600, which slightly exceeds the initial rating of 1500), \textit{Advanced} (Rapid rating between 1600 and 2000), and \textit{Master} (Rapid rating over 2000).
% , roughly comprising the top 10\% of players\cite{lichess_rapid}).
In addition, we aim to evaluate move prediction across diverse skill combinations
% , which \textit{Maia Testset} excludes. Therefore, we construct 
with the \textit{Cross-skill Testset} constructed from Dec 2023 games.
% a new testing dataset from Lichess games played in December 2023, 
% ensuring that positions under each skill level combination have sufficient data to be statistically reliable. 
Finally, we construct \textit{Grounded Testset} with 450,000 positions that has recorded Stockfish evaluations, 
% from December 2023 games in \href{https://database.lichess.org/\#evals}{Lichess Database} 
% where Stockfish evaluations are available. 
% Such recorded evaluations 
which can serve as grounded facts to measure move quality. %, centipawn loss, win-rate loss, and distinguish blunders.
Statistics of datasets are summarized in Appendix Table~\ref{tab:maia2data_stats}.

\begin{table}[!t]
  \centering
  \small
  % \vspace{-0.4cm}
  % \renewcommand\arraystretch{0.8}
  \caption{Move prediction accuracy on the \textit{Maia-2 Testset}. \textit{Skilled}, \textit{Advanced}, and \textit{Master} are grouped according to Section~\ref{section:testset} and \textit{Avg} denotes macro-averaged results.}
  % \vspace{-0.2cm}
  \setlength{\tabcolsep}{3.5pt}
    \begin{tabular}{lccc|ccc|ccc|cc}
    \toprule
          & \multicolumn{3}{c|}{Stockfish} & \multicolumn{3}{c|}{Leela} & \multicolumn{3}{c|}{Maia} & \multirow{2.5}{*}{Maia-2$_{\text{subset}}$} & \multirow{2.5}{*}{Maia-2} \\
          \cmidrule(lr){2-4} \cmidrule(lr){5-7} \cmidrule(lr){8-10} %\cmidrule(lr){11-12}
          & 3 & 9 & 15 & 1500 & 2200 & 3200 & 1100 & 1500 & 1900 & & \\
    \midrule
    Skilled & 36.22 & 36.00 & 36.86 & 40.46 & 39.79 & 39.97 & 51.48 & 50.79 & 48.51 & 51.51 & \textbf{51.72} \\
    Advanced & 38.25 & 38.78 & 39.83 & 44.45 & 43.97 & 44.29 & 49.13 & 52.61 & 52.26 & 53.54 & \textbf{54.15} \\
    Master & 40.71 & 43.26 & 44.61 & 48.69 & 47.11 & 47.75 & 45.85 & 50.76 & 53.20 & 53.16 & \textbf{53.87} \\
    Avg & 38.39 & 39.35 & 40.43 & 44.53 & 43.62 & 44.00 & 48.82 & 51.39 & 51.32 & 52.74 & \textbf{53.25} \\
    \bottomrule
    \end{tabular}%
  \label{tab:baseline}%
  % \vspace{-0.4cm}
\end{table}%

% \begin{table}[!t]
%   \centering
%       % \renewcommand\arraystretch{0.9}
%   	% \renewcommand\tabcolsep{2pt}
%   \caption{Move prediction accuracy on the \textit{Maia-1 Testset}. \textit{Skilled}, \textit{Advanced}, and \textit{Master} are grouped according to Section~\ref{section:testset} and \textit{Avg} denotes macro-averaged results.}
%   % \vspace{-0.3cm}
%     \begin{tabular}{lcccc}
%     \toprule
%           & Skilled & Advanced & Master & Avg \\
%         \cmidrule(l){2-4} \cmidrule(lr){5-5}
%     Stockfish 3 & 36.22 & 38.25 & 40.71 & 38.39 \\
%     Stockfish 7 & 35.66 & 38.08 & 42.25 & 38.66 \\
%     Stockfish 9 & 36.00 & 38.78 & 43.26 & 39.35 \\
%     Stockfish 11 & 36.38 & 39.33 & 43.71 & 39.81 \\
%     Stockfish 15 & 36.86 & 39.83 & 44.61 & 40.43 \\
%     \cmidrule(l){2-4} \cmidrule(lr){5-5}
%     Leela 1500 & 40.46 & 44.45 & 48.69 & 44.53 \\
%     Leela 2200 & 39.79 & 43.97 & 47.11 & 43.62 \\
%     Leela 3200 & 39.97 & 44.29 & 47.75 & 44.00 \\
%     Leela 3700 & 40.47 & 44.65 & 48.12 & 44.41 \\
%     \cmidrule(l){2-4} \cmidrule(lr){5-5}
%     Maia 1100 & 51.48 & 49.13 & 45.85 & 48.82 \\
%     Maia 1500 & 50.79 & 52.61 & 50.76 & 51.39 \\
%     Maia 1900 & 48.51 & 52.26 & 53.20 & 51.32 \\
%     \cmidrule(l){2-4} \cmidrule(lr){5-5}
%     Maia-2$_{\text{subset}}$ & 51.51 & 53.54 & 53.16 & 52.74 \\
%     Maia-2 & \textbf{51.72} & \textbf{54.15} & \textbf{53.87} & \textbf{53.25} \\
%     \bottomrule
%     \end{tabular}%
%   \label{tab:baseline}%
%   % \vspace{-0.4cm}
%   % Table generated by Excel2LaTeX from sheet ' Move Matching'
% \end{table}%

\subsection{Move Prediction Accuracy}

\xhdr{Maia-2} In Table~\ref{tab:baseline}, we show the top-1 move prediction accuracy of all models across all groups of players on the \textit{Maia-1 Testset}. Maia-2 demonstrates strong and consistent performance across all skill levels, surpassing all baselines. Specifically, despite Maia-1 models being specifically trained to mimic chess moves by players at specific skill levels, Maia-2 emerges as a unified one-for-all model that is consistently effective across the entire spectrum of chess skills.
The largest improvement is on \emph{Advanced} players, where Maia-2 gains 1.5 percentage points over the nearest competitor (Maia 1500). When averaging across skill levels, Maia-2 outperforms all other models by almost 2 full percentage points in overall accuracy. 
Note that the ceiling accuracy of human move prediction is far below 100\% given the randomness and diversity of human decisions—even the \emph{same} player won’t always make the same decision when faced with the same position. Our 2 percentage point gain is substantial considering that the difference between Maia-1 and Leela, the previous state-of-the-art model for this task and a traditional chess engine not trained for this task at all, is only 6 percentage points.
Furthermore, Maia-1 is essentially a mixture of 9 experts targeting the specific players’ skill level, where each expert has 10.3M parameters. Regarding the routing function to select the best-performing expert as a nonparameterized function, Maia-1 has 92M parameters in total. Maia-2, on the other hand, is a one-for-all model with 23.3M parameters under our default settings. Therefore, Maia-2 achieves better human move prediction accuracy with even much fewer trainable parameters.

\xhdr{Baseline models} Both Maia-2 and Maia-1 significantly outperform Stockfish and Leela, typically by 5--15 percentage points. Note that Stockfish and Leela aim to play optimal chess (as most humans do too), and only ``predict'' human moves when their approximations to optimality happen to overlap with those of human players. However, we compare to these traditional chess engines because besides Maia-1, there are still the default method of creating ``human-like'' AI agents. The accuracy gap between Maia-1 architectures and traditional chess engines demonstrates the necessity of developing specialized models to mimic human chess moves.

\begin{figure*}[!t]
\vspace{-0.4cm}
	\centering
\includegraphics[width=0.95\textwidth]{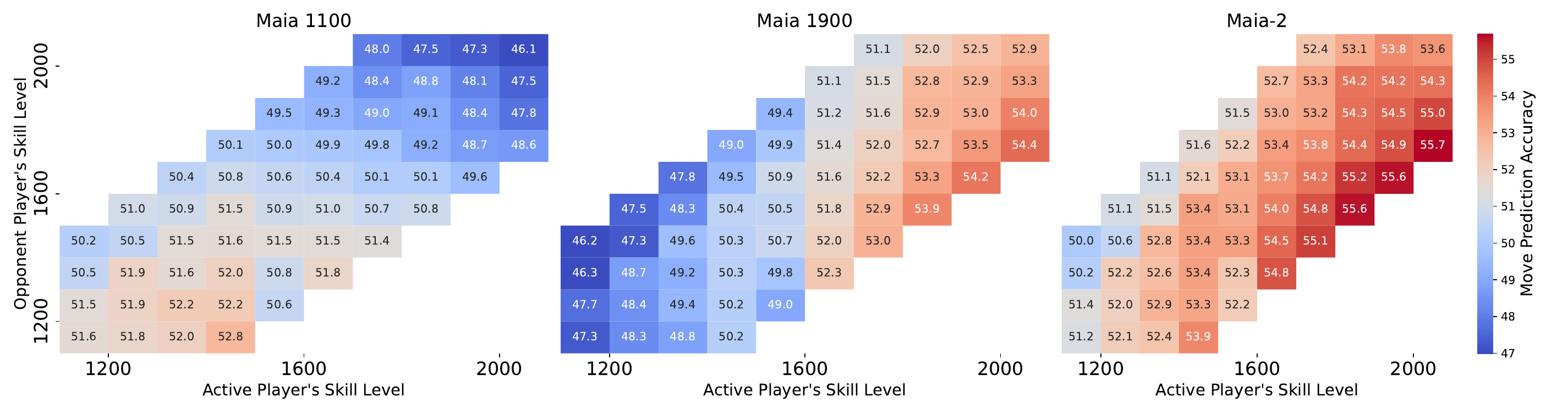}
 \vspace{-0.4cm}
        \caption{Move prediction accuracy across diverse skill levels. Colors represent performance, with warmer tones indicating higher accuracy.}
  \label{fig:heatmap}
\vspace{-0.4cm}
\end{figure*}

\xhdr{Maia-2$_{\text{subset}}$} Maia-2 differs from Maia-1 in two main ways: it has a different architecture and it has access to more training data. To control for the difference in training data and isolate the effects of our architecture, we create Maia-2$_{\text{subset}}$ which has access to the exact same training data that Maia-1 was developed with. Comparing the two, we see that Maia-2$_{\text{subset}}$ matches or outperforms all baselines and alternate models. 
Recall that Maia-2 and Maia-2$_{\text{subset}}$ don't have the recent history passed as input to them, yet still achieve state-of-the-art results. 
It is important to note that each Maia-1 model is specifically trained for its respective skill level, relying solely on games where the active and opponent skill levels match for its training data. On the contrary, the unified modeling approach with skill-aware attention of Maia-2$_{\text{subset}}$ allows it to utilize a broader spectrum of games, featuring a variety of skill-level pairings, for training purposes. Consequently, while both Maia-1 and Maia-2$_{\text{subset}}$ draw from the same source dataset, Maia-2$_{\text{subset}}$ can leverage a significantly larger portion of this data for its training, improving its learning and predictive capabilities.
The improvement from Maia-2$_{\text{subset}}$ to Maia-2 underscores the importance of extensive training with vast datasets. A broader range of games provides Maia-2 with access to more comprehensive and nuanced patterns in human chess moves. Using Maia-2$_{\text{subset}}$ as a comparison, we can determine the relative contributions of model architecture and training data to Maia-2's 1.9 percentage point gap over its nearest rival (Maia 1500). This calculation suggests that 73\% of the increase in performance is due to the architecture improvements and 27\% is due to increased training data. 

\xhdr{Move prediction perplexity}
While top-1 accuracy gains are important, they may overshadow larger improvements in prediction quality. To account for this, we also measure the perplexity of move predictions
%The perplexity of move predictions is employed as an additional metric accompanying the move prediction accuracy to evaluate how well a model predicts human moves
, which reflects the model's confidence in its predictions. A lower perplexity indicates the model is more confident and accurate in human move prediction, as it corresponds to a higher likelihood of the correct human move.
As shown in Table~\ref{tab:perplexity}, Maia-2 consistently yields substantially lower perplexity in all groups of skill levels compared to Maia-1. In particular, Maia-2 significantly outperforms Maia-1 in \textit{Advanced} and \textit{Master} moves with relatively large margins, demonstrating the effectiveness of our unified modeling approach across diverse skill levels.

\begin{table}[t]
  \centering
  \vspace{-0.4cm}
  \begin{tabular}{cc}
    \begin{minipage}[t]{0.48\textwidth}
      \centering
      \caption{Move prediction perplexity on the \textit{Grounded Testset}.}
      \resizebox{\textwidth}{!}{
        \begin{tabular}{lcccc}
          \toprule
                & Skilled & Advanced & Master & Avg \\
          \cmidrule(l){2-4} \cmidrule(lr){5-5}
          Maia 1100 & 5.05 & 5.56 & 6.01 & 5.54 \\
          Maia 1500 & 4.98 & 5.31 & 5.50 & 5.26 \\
          Maia 1900 & 4.44 & 4.71 & 4.86 & 4.67 \\
          \cmidrule(l){2-4} \cmidrule(lr){5-5}
          \modelname{} & \textbf{4.30} & \textbf{3.90} & \textbf{4.02} & \textbf{4.07} \\
          \bottomrule
        \end{tabular}
      }
      \label{tab:perplexity}
    \end{minipage}
    &
    \begin{minipage}[t]{0.48\textwidth}
      \centering
      \caption{Ablation study results, where \modelname{}$_{\text{subset}}$ is compared with versions without skill-aware attention (``w/o Att'') and without infusing auxiliary information (``w/o Aux'').}
      \resizebox{\textwidth}{!}{
        \begin{tabular}{lcccc}
          \toprule
                & Skilled & Advanced & Master & Avg \\
          \cmidrule(l){2-4} \cmidrule(lr){5-5}
          w/o Att & 50.63 & 52.89 & 51.73 & 51.75 \\
          w/o Aux & 50.96 & 52.98 & 52.34 & 52.09 \\
          \cmidrule(l){2-4} \cmidrule(lr){5-5}
          \modelname{}$_{\text{subset}}$ & \textbf{51.51} & \textbf{53.54} & \textbf{53.16} & \textbf{52.74} \\
          \bottomrule
        \end{tabular}
      }
      \label{tab:ablation}
    \end{minipage}
    \vspace{-0.3cm}
  \end{tabular}
\end{table}

% \begin{table}[!t]
%   \centering
%         % \renewcommand\arraystretch{0.7}
%   	% \renewcommand\tabcolsep{2pt}
%   \caption{Move prediction perplexity on the \textit{Grounded Testset}.}
%   % \vspace{-0.3cm}
%     \begin{tabular}{lcccc}
%     \toprule
%           & Skilled & Advanced & Master & Avg \\
%         \cmidrule(l){2-4} \cmidrule(lr){5-5}
%     Maia 1100 & 5.05 & 5.56 & 6.01 & 5.54 \\
%     Maia 1500 & 4.98 & 5.31 & 5.50 & 5.26 \\
%     Maia 1900 & 4.44 & 4.71 & 4.86 & 4.67 \\
%     \cmidrule(l){2-4} \cmidrule(lr){5-5}
%     Maia-2 & \textbf{4.30} & \textbf{3.90} & \textbf{4.02} & \textbf{4.07}\\
%     \bottomrule
%     \end{tabular}%
%   \label{tab:perplexity}%
% \end{table}%

\xhdr{Adaptive move predictions}
We now evaluate Maia-2's ability to predict compare across diverse skill combinations using the \textit{Cross-skill Testset}.
As shown in Figure~\ref{fig:heatmap}, Maia-2 consistently outperforms both Maia 1100 and Maia 1900 in almost all combinations of active player and opponent player skill levels.
In particular, although Maia 1100 and Maia 1900 demonstrate competent performance within their respective domains of expertise, their predictive accuracy decreases substantially outside of these targeted skill levels. This is because Maia-1 models are static and cannot respond to varied skill levels and adjust their predictions accordingly. 
In contrast, our proposed unified modeling approach with skill-aware attention enables Maia-2 to adapt its predictions to account for the skill levels of both the active player and the opponent player, so that varying skill level configurations correspondingly can result in better aligned human move predictions. More results on comparisons with other Maia-1 versions can be found in Figure~\ref{fig:heatmap_appendix} in the Appendix.

\xhdr{Move quality} One of our key motivations for creating a unified model of human chess behavior is to guide the development of future algorithmic learning tools. As such, understanding the mistakes that people make is of fundamental interest. Can Maia-2 predict mistakes better than Maia-1? Figure~\ref{fig:acc_winprob} in the Appendix shows the move prediction accuracy on the \textit{Grounded Testset} as a function of move quality, measured by win-rate loss, which is calculated following the same procedures as prior studies~\cite{mcilroy2020aligning, mcilroy2022learning}. All models generally decrease in their ability to predict worse moves, since humans are generally trying to avoid mistakes, and high-quality moves are more certain whereas lower-quality moves can be more random and thus hard to predict.
Nevertheless, Maia-2 outperforms all versions of Maia-1 across most of the move quality range, demonstrating the effectiveness of our unified modeling approach for human move prediction. %Maia-2's overall gains are not constrained to any specific move quality type, but are spread across the entire range.

\begin{figure}[!t]
% \vspace{-0.4cm}
	\centering
	\includegraphics[width=1.02\textwidth]{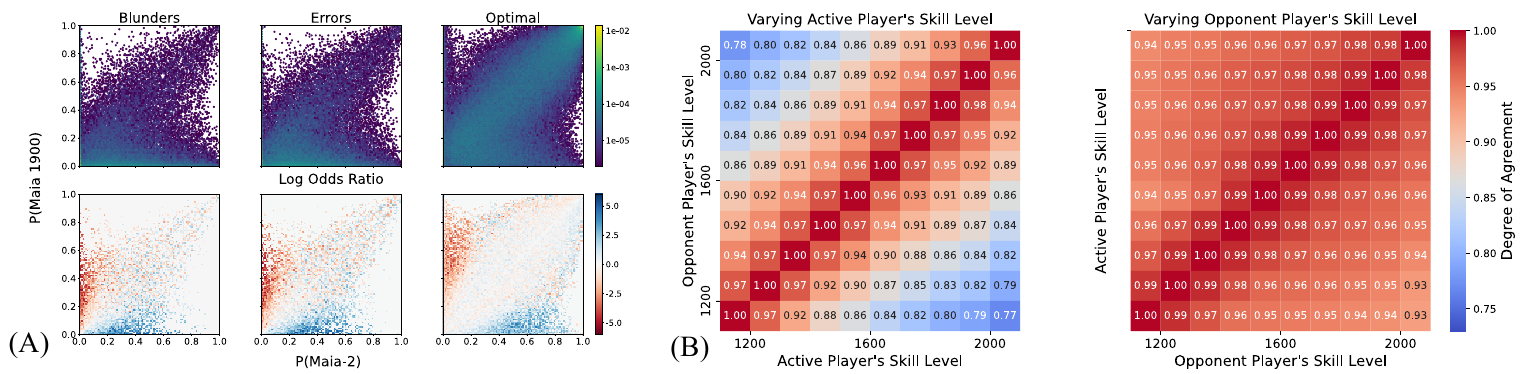}
  % \vspace{-0.4cm}
        \caption{(A). (Top) Joint probability assigned to human moves played by \modelname{} ($x$) and Maia 1900 ($y$), split by move quality. Blunders (left) reduce the expected win-rate by $\geq$ 10\%, Errors (middle) by 5--10\%, and Optimal  (right) by $\leq 0\%$. (Bottom) Log odds ratio of $p(x,y)$ and $p(y, x)$ from top. (B). Move prediction agreement as (left) active player and (right) opponent player skill are varied.
        All cells are evaluated on the same set of positions but with altered skill level configurations. 
       }
  \label{fig:joint_prob}

\end{figure}

We are also interested in how certain the models are about their predictions of various move qualities. 
Figure~\ref{fig:joint_prob}.(A)(top) shows the probabilities that Maia-2 (x-axis) and Maia-1 (y-axis) attribute to the moves people actually played in \textit{Grounded Testset}. 
%, indicated by the color gradient, shows the frequency of joint probability occurrences on the .
The concentration of points in the lower right quadrant (closer to P(Maia-2) = 1 and P(Maia 1900) = 0) suggests that Maia-2 assigns a higher probability to the true move than Maia-1 does, indicating superior predictive performance.
Conversely, the less dense upper left quadrant indicates fewer instances where Maia 1900 outperforms Maia-2. 
%Figure~\ref{fig:joint_prob} shows that Maia-2 is more confident and accurate in predicting the correct moves, as evidenced by the greater density of points towards the bottom right, signifying better alignment with actual human moves. 
Remarkably, while this consistently occurs across all move qualities, the distinction is more pronounced for \textit{Blunders} and \textit{Errors} compared to \textit{Optimal} moves. %This pattern can be attributed to the design of Maia 1900, which is optimized for higher skill-level play, thus favoring the prediction of higher-quality moves and deteriorating drastically in predicting lower-quality moves. 
Additionally, the bottom row of Figure \ref{fig:joint_prob}.(A) shows the log odds ratio between $P(x,y)$ and $P(y,x)$ in the top row. The abundance of blue points below the diagonal indicates that Maia-2 is almost always more confident in the correct move than Maia-1 is, indicating an across-the-board improvement in move prediction. Maia-2 offers superior and more confident prediction across diverse move qualities. 
We also conduct an ablation study as shown in Table~\ref{tab:ablation}, we point the readers to Appendix~\ref{appendix:more_results} for more information.

% \begin{figure}[!t]
% % \vspace{-0.4cm}
% 	\centering
% 	\includegraphics[width=0.85\textwidth]{Figs/joint_prob_plot.pdf}
%   % \vspace{-0.4cm}
%         \caption{(Top) Joint probability assigned to  human moves played by Maia-2 ($x$) and Maia 1900 ($y$), split by move quality. Blunders (left) reduce the expected win-rate by $\geq$ 10\%, Errors (middle) by 5--10\%, and Optimal  (right) by $\leq 0\%$. (Bottom) Log odds ratio of $p(x,y)$ and $p(y, x)$ from top.
%        }
%   \label{fig:joint_prob}
%  % \vspace{-0.5cm}
% \end{figure}

\subsection{Move Prediction Coherence}
Maia-2's accuracy across the spectrum of human skill is certainly desirable, but perhaps an even more important dimension is prediction \emph{coherence} as skill varies. A central drawback of Maia-1 is that it models players at different skill levels independently from each other, which results in particularly volatile predictions: the same position might elicit very different predicted behavior from models of adjacent skill levels. This is problematic because we know from personal experience that this type of volatility is rare: players don't change that much as they improve. This limits Maia's ability to perform well in downstream tasks such as serving as a teaching aid, as its understanding of one skill level bears little resemblance to its understanding of the next. In reality, players move from one skill level to another by making small, consistent adjustments. Does Maia-2 reflect this behavioral coherence?

\xhdr{Prediction smoothness} 
We measure the coherence of Maia-2's predictions by testing for smoothness features in its entire set of predictions. 
Call a model's treatment of a position \emph{monotonic} if the predicted probability of the correct move increases with skill monotonically. In the \textit{Grounded Testset} of 100K positions, we find that Maia-1 only treats 1\% of them monotonically. In stark contrast, however, Maia-2 treats 27\% of them monotonically, clearly demonstrating that Maia-2 is much more coherent. 
Similarly, call a model's treatment of a position \emph{transitional} if it predicts a suboptimal move for some prefix of skills and then transitions to an optimal move for all subsequent skill levels. Again, Maia-2 treats substantially more positions transitionally---around 22\% of them compared with 17\% for Maia. 

It’s important to note that Maia-2 is deliberately designed to encourage coherence across skill levels without rigidly enforcing it. Our objective is not to impose coherence as a hard constraint, which might obscure legitimate differences in player behavior between skill levels, but to create a model architecture that naturally encourages coherence where the data supports it.

\begin{table}[!t]
\small
    \renewcommand\arraystretch{0.7}
  \centering
  \caption{Percentage of monotonic and transitional positions.}
   \vspace{-0.2cm}
    \begin{tabular}{ccccccc}
    \toprule
          & \multicolumn{3}{c}{\%Monotonic} & \multicolumn{3}{c}{\%Transitional} \\
\cmidrule(lr){2-4} \cmidrule(lr){5-7}          & Skilled & Advanced & Master & Skilled & Advanced & Master \\
\cmidrule{2-7}    Maia-1 & 1.61  & 1.42  & 1.14  & 13.34 & 18.14 & 20.48 \\
    Maia-2 & 27.61 & 28.51 & 26.38 & 22.59 & 23.39 & 21.72 \\
    \bottomrule
    \end{tabular}%
  \label{tab:smoothness}%
  \vspace{-0.2cm}
\end{table}%

\xhdr{Move prediction agreement}
% \begin{figure}[!t]
% 	\centering
% 	\includegraphics[width=0.85\textwidth]{Figs/agreement.pdf}
%  % \vspace{-0.2cm}
%         \caption{Move prediction agreement as (left) active player and (right) opponent player skill are varied.  %with the left matrix for fixed opponent player skill across varying active skill levels, and the right matrix for fixed active skill level across varying opponent player skill levels. 
%         All cells are evaluated on the same set of positions but with altered skill level configurations. 
%         The value in each cell denotes the percentage of predictions that agree with the diagonal of the same row, where active and opponent skill levels are set equal.}
%   \label{fig:agreement}
% % \vspace{-0.4cm}
% \end{figure}
As a first test, we measure move prediction coherence as we vary active player skill and opponent player skill in Maia-2. The results shown in Figure~\ref{fig:joint_prob}.(B) reveal several trends. First, increasingly varying either the active or opponent rating results in lower agreement, suggesting that Maia-2 smoothly varies its predictions with skill. 
%Such coherent performance is a direct benefit of our unified modeling approach that involves and effectively models chess positions with diverse pairs of active and opponent skill levels. 
%In particular, our data balancing strategy could be a crucial component as it ensures sufficient and balanced training data within each active--opponent skill level pair.
Second, comparing the two heatmaps reveals that Maia-2 has clearly learned that varying one's own skill has much larger effects than varying the opponent's---changing one's own skill against a fixed opponent can change the decision up to 22\% of the time, but changing the opponent's skill while fixing our own skill will only change the decision up to 6\% of the time. This is intuitive, as players must change their decisions in order to play at a higher level, while in theory one's opponent shouldn't affect one's decision. Of course, humans are not optimal agents and sometimes take their opponent's skill level into account when deciding on a move---willfully or not---which is reflected in our results. 

\begin{figure}[t]
% \vspace{-0.3cm}
	\centering
	\includegraphics[width=0.95\textwidth]{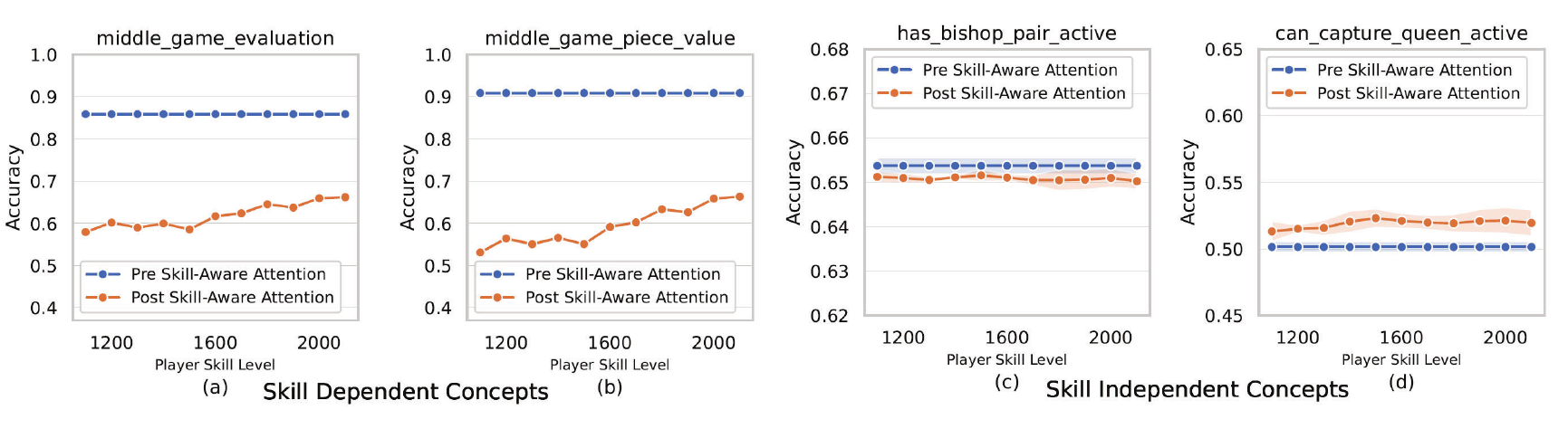}
  % \vspace{-0.3cm}
        \caption{\modelname{}'s chess concept recognition as a function of skill level, as measured by linear activation probes right before (blue) and after (orange) skill-aware attention. (a) Stockfish overall board evaluation for middle-game positions. (b) Stockfish evaluation of middle-game bonuses and penalties to pieces for white. (c) Does the active player own two bishops? (d) Can the active player capture the opponent's queen? }
  \label{fig:probing}
 % \vspace{-0.4cm}
\end{figure}

\xhdr{Chess concept understanding}
Human chess players of varying strengths differ in their ability to recognize important features and patterns on the board, e.g., stronger players are adept at discerning subtle nuances.
% , including accurately evaluating the state of the board and understanding the strengths and weaknesses of their position, that often elude their weaker counterparts
% . While we've already shown that \modelname{} has a remarkable capability to adapt its move predictions according to different skill levels, 
We now turn our focus to a critical question: does \modelname{} vary in its ability to capture  human chess concepts when given different skill levels? 
%This investigation offers a  window into the interpretability of \modelname{}, further bridging the gap between super-human AI and human-understandable chess knowledge
Following the chess concepts probing strategy for AlphaZero~\cite{mcgrath2022acquisition}, 
% we train linear probes on internal representations before and after the skill-aware attention blocks of \modelname{}, and use their test performance as a proxy for capturing chess concepts. In Figure~\ref{fig:probing}, 
we show how \modelname{}'s grasp of various concepts varies with skill. The left two plots in Figure~\ref{fig:probing} show concepts for which \modelname{} clearly distinguishes between skill levels, with higher-skill players paying more attention to them than lower-skill players. These are general board evaluations as given by \href{https://hxim.github.io/Stockfish-Evaluation-Guide}{Stockfish}~\cite{stockfish}, or aggregate piece values. Note that pre-skill-aware attention is always flat because by construction it cannot vary with skill, since skill-aware attention has not been applied yet. The two plots on the right depict concepts that live closer to fundamental chess rules, and as such are less dependent on player skill. For skill-dependent concepts, the figures reveal an increasing trend in mastery level after skill-aware attention, aligning with the increase in dedicated skill levels. Meanwhile, the model's mastery level decreases after passing through the skill-aware attention modules, potentially adjusting for the imperfections of human players. Conversely, the skill-aware attention blocks are not responsive to skill-independent concepts.

\section{Discussion}

\xhdr{Human Study} 
% In an ideal human experiment, we would give a position to a human at a particular rating, and compare their chosen move to our model output. Our experiments do exactly this, and thus we view them as massive human studies that measure the move-matching accuracy and coherence with the recorded behaviors of real humans. 
In addition to human move matching, we also consider engagement, another dimension of human study. In particular, we implement a randomized experiment on Lichess: human players challenge our bots, and we randomize whether players play against Maia-1 or Maia-2. Our result is that our higher move-matching and our vastly improved coherence, across all skill levels, come at no cost to human subject engagement, and in fact slightly increase engagement: players rematch Maia-2 almost 1 percentage point more than Maia-1 (41.2\% vs. 40.3\%). Although engagement is not our main objective, this is further promising evidence that we have achieved a more human-aligned model that coherently captures human style across different skill levels.

% \xhdr{A Chess Foundational Model} Maia-2 is a unified model architecture that can accurately and coherently capture human decision-making in chess across a broad spectrum of skill levels. 
% As such, Maia-2 can be regarded as a foundation model for human-AI alignment 
% in chess, serving as a base model for general-purpose explorations of human-like AI.
% For example, fine-tuning Maia-2 with personal historical data of increasing active player skill levels and diverse opponent player skill levels can yield personalized move predictions, facilitating algorithmic education tools and more relatable AI partners. 
% Also, the prior knowledge and general understanding acquired during pre-training can help Maia-2 to be data efficient when extending to grandmaster level moves, which are relatively rare and thus cannot be learned with the previous skill-independent modeling approaches.

\xhdr{Ethical Considerations} 
% The development of human-like models raises ethical concerns discussed in previous work~\cite{mcilroy2022learning,mcilroy2022mimetic}. 
We believe Maia-2 poses limited risk while offering large potential benefits. 
Our data is highly aggregated, with almost 1 billion games being used for training, and chess as a domain is generally low-risk. 
Meanwhile, helping people improve in chess could lead to increased cognitive skills, confidence boosts, and help with general life satisfaction. 
Our vision is for Maia-2 to power AI partners and training aids; it cannot currently replace skilled human tutors and coaches.

\xhdr{Limitation} Our work has limitations. First, we are excited by the applications that Maia-2 will enable, such as more relatable AI partners and AI-powered learning aids, the development of which is out of scope for the current work. Maia-2 does not yet incorporate search, although previous work has demonstrated that with proper regularization it can help improve move prediction performance~\cite{jacob2022modeling}. Relatedly, we group the strongest players in a single bucket, although modeling the very best players in the world remains difficult due to the complexity and depth of their moves.

% \begin{itemize}
%     \item Better modeling of humans raises ethical concerns as discussed in previous works~\cite{mcilroy2022learning,mcilroy2022mimetic}
%     \item The creation of large models that can parrot what they have been trained on is an area of increasing concern by the people who generate the data
%     \item We believe that these models will be of significant benefit to both the scientific community and the chess community (Lichess) that we collected the data from
%     \item Lichess has had a very positive reaction to the previous models
%     \item in theory our models could be imitating specific `close to model players' more precisely than others, meaning that practice on our models by other players would disadvantage the `close to model players' over other players on the site
%     \item We have no evidence of this, but by using our model as a base model this is possible
%     \item These models also might reduce the `humanness' of chess, by reducing the need for tutors and pushing players closer to the types of games our models lead to
%     \item Both of these concerns are only theoretical at this point and we hope that they do not come to pass
% \end{itemize}

\clearpage
\section*{Acknowledgements}
We thank the anonymous reviewers for helpful comments. 
This research was supported in part by a grant from the Natural Sciences and Engineering Research Council of Canada (NSERC), a Microsoft Research Award, a Simons Collaboration grant, and a grant from the MacArthur Foundation.

\clearpage
\bibliographystyle{unsrt}
\bibliography{09-main-cites}

\begin{thebibliography}{10}

\bibitem{mcilroy2020aligning}
Reid McIlroy-Young, Siddhartha Sen, Jon Kleinberg, and Ashton Anderson.
\newblock Aligning superhuman ai with human behavior: Chess as a model system.
\newblock In {\em Proceedings of the 26th ACM SIGKDD International Conference on Knowledge Discovery \& Data Mining}, pages 1677--1687, 2020.

\bibitem{hsu1999ibm}
Feng-hsiung Hsu.
\newblock Ibm's deep blue chess grandmaster chips.
\newblock {\em IEEE micro}, 19(2):70--81, 1999.

\bibitem{clark1999sciences}
William Clark, Jan Golinski, and Simon Schaffer.
\newblock {\em The sciences in enlightened Europe}.
\newblock University of Chicago Press, 1999.

\bibitem{turochamp}
Frederic Friedel.
\newblock Reconstructing turing's "paper machine".

\bibitem{shannon2001mathematical}
Claude~E. Shannon.
\newblock A mathematical theory of communication.
\newblock {\em Bell Syst. Tech. J.}, 27:623--656, 1948.

\bibitem{stockfish}
Tord Romstad, Marco Costalba, Joona Kiiski, and et~al.
\newblock Stockfish.
\newblock stockfishchess.org, 2023.
\newblock Accessed: 2024-01-05.

\bibitem{silver2016mastering}
David Silver, Aja Huang, Chris~J Maddison, Arthur Guez, Laurent Sifre, George Van Den~Driessche, Julian Schrittwieser, Ioannis Antonoglou, Veda Panneershelvam, Marc Lanctot, et~al.
\newblock Mastering the game of go with deep neural networks and tree search.
\newblock {\em nature}, 529(7587):484--489, 2016.

\bibitem{feng2023chessgpt}
Xidong Feng, Yicheng Luo, Ziyan Wang, Hongrui Tang, Mengyue Yang, Kun Shao, David Mguni, Yali Du, and Jun Wang.
\newblock Chessgpt: Bridging policy learning and language modeling.
\newblock {\em arXiv preprint arXiv:2306.09200}, 2023.

\bibitem{hsu2002behind}
Feng-Hsiung Hsu.
\newblock {\em Behind Deep Blue: Building the computer that defeated the world chess champion}.
\newblock Princeton University Press, 2002.

\bibitem{silver2017mastering}
David Silver, Thomas Hubert, Julian Schrittwieser, Ioannis Antonoglou, Matthew Lai, Arthur Guez, Marc Lanctot, Laurent Sifre, Dharshan Kumaran, Thore Graepel, et~al.
\newblock Mastering chess and shogi by self-play with a general reinforcement learning algorithm.
\newblock {\em arXiv}, 2017.

\bibitem{zahavy2023diversifying}
Tom Zahavy, Vivek Veeriah, Shaobo Hou, Kevin Waugh, Matthew Lai, Edouard Leurent, Nenad Tomasev, Lisa Schut, Demis Hassabis, and Satinder Singh.
\newblock Diversifying ai: Towards creative chess with alphazero.
\newblock {\em arXiv preprint arXiv:2308.09175}, 2023.

\bibitem{vaswani2017attention}
Ashish Vaswani, Noam Shazeer, Niki Parmar, Jakob Uszkoreit, Llion Jones, Aidan~N Gomez, {\L}ukasz Kaiser, and Illia Polosukhin.
\newblock Attention is all you need.
\newblock {\em Advances in neural information processing systems}, 30, 2017.

\bibitem{brown2020language}
Tom Brown, Benjamin Mann, Nick Ryder, Melanie Subbiah, Jared~D Kaplan, Prafulla Dhariwal, Arvind Neelakantan, Pranav Shyam, Girish Sastry, Amanda Askell, et~al.
\newblock Language models are few-shot learners.
\newblock {\em Advances in neural information processing systems}, 33:1877--1901, 2020.

\bibitem{touvron2023llama}
Hugo Touvron, Louis Martin, Kevin Stone, Peter Albert, Amjad Almahairi, Yasmine Babaei, Nikolay Bashlykov, Soumya Batra, Prajjwal Bhargava, Shruti Bhosale, et~al.
\newblock Llama 2: Open foundation and fine-tuned chat models.
\newblock {\em arXiv preprint arXiv:2307.09288}, 2023.

\bibitem{kojima2022large}
Takeshi Kojima, Shixiang~Shane Gu, Machel Reid, Yutaka Matsuo, and Yusuke Iwasawa.
\newblock Large language models are zero-shot reasoners.
\newblock {\em Advances in neural information processing systems}, 35:22199--22213, 2022.

\bibitem{wu2023autogen}
Qingyun Wu, Gagan Bansal, Jieyu Zhang, Yiran Wu, Shaokun Zhang, Erkang Zhu, Beibin Li, Li~Jiang, Xiaoyun Zhang, and Chi Wang.
\newblock Autogen: Enabling next-gen llm applications via multi-agent conversation framework.
\newblock {\em arXiv preprint arXiv:2308.08155}, 2023.

\bibitem{noauthor_llmchess_nodate}
Max Hager.
\newblock {LLMChess}.

\bibitem{carlini_carlinichess-llm_2024}
Nicholas Carlini.
\newblock carlini/chess-llm.

\bibitem{srivastava2022beyond}
Aarohi Srivastava, Abhinav Rastogi, Abhishek Rao, Abu Awal~Md Shoeb, Abubakar Abid, Adam Fisch, Adam~R Brown, Adam Santoro, Aditya Gupta, Adri{\`a} Garriga-Alonso, et~al.
\newblock Beyond the imitation game: Quantifying and extrapolating the capabilities of language models.
\newblock {\em arXiv preprint arXiv:2206.04615}, 2022.

\bibitem{lee2022improving}
Andrew Lee, David Wu, Emily Dinan, and Mike Lewis.
\newblock Improving chess commentaries by combining language models with symbolic reasoning engines.
\newblock {\em arXiv preprint arXiv:2212.08195}, 2022.

\bibitem{noever2020chess}
David Noever, Matt Ciolino, and Josh Kalin.
\newblock The chess transformer: Mastering play using generative language models.
\newblock {\em arXiv preprint arXiv:2008.04057}, 2020.

\bibitem{mcgrath2022acquisition}
Thomas McGrath, Andrei Kapishnikov, Nenad Toma{\v{s}}ev, Adam Pearce, Martin Wattenberg, Demis Hassabis, Been Kim, Ulrich Paquet, and Vladimir Kramnik.
\newblock Acquisition of chess knowledge in alphazero.
\newblock {\em Proceedings of the National Academy of Sciences}, 119(47):e2206625119, 2022.

\bibitem{schut2023bridging}
Lisa Schut, Nenad Tomasev, Tom McGrath, Demis Hassabis, Ulrich Paquet, and Been Kim.
\newblock Bridging the human-ai knowledge gap: Concept discovery and transfer in alphazero.
\newblock {\em arXiv preprint arXiv:2310.16410}, 2023.

\bibitem{mcilroy2021detecting}
Reid McIlroy-Young, Yu~Wang, Siddhartha Sen, Jon Kleinberg, and Ashton Anderson.
\newblock Detecting individual decision-making style: Exploring behavioral stylometry in chess.
\newblock {\em Advances in Neural Information Processing Systems}, 34:24482--24497, 2021.

\bibitem{mcilroy2022learning}
Reid McIlroy-Young, Russell Wang, Siddhartha Sen, Jon Kleinberg, and Ashton Anderson.
\newblock Learning models of individual behavior in chess.
\newblock In {\em Proceedings of the 28th ACM SIGKDD Conference on Knowledge Discovery and Data Mining}, pages 1253--1263, 2022.

\bibitem{jacob2022modeling}
Athul~Paul Jacob, David~J Wu, Gabriele Farina, Adam Lerer, Hengyuan Hu, Anton Bakhtin, Jacob Andreas, and Noam Brown.
\newblock Modeling strong and human-like gameplay with kl-regularized search.
\newblock In {\em International Conference on Machine Learning}, pages 9695--9728. PMLR, 2022.

\bibitem{littman1994markov}
Michael~L Littman.
\newblock Markov games as a framework for multi-agent reinforcement learning.
\newblock In {\em Machine learning proceedings 1994}, pages 157--163. Elsevier, 1994.

\bibitem{he2016deep}
Kaiming He, Xiangyu Zhang, Shaoqing Ren, and Jian Sun.
\newblock Deep residual learning for image recognition.
\newblock In {\em Proceedings of the IEEE conference on computer vision and pattern recognition}, pages 770--778, 2016.

\bibitem{dosovitskiy2020image}
Alexey Dosovitskiy, Lucas Beyer, Alexander Kolesnikov, Dirk Weissenborn, Xiaohua Zhai, Thomas Unterthiner, Mostafa Dehghani, Matthias Minderer, Georg Heigold, Sylvain Gelly, et~al.
\newblock An image is worth 16x16 words: Transformers for image recognition at scale.
\newblock {\em arXiv preprint arXiv:2010.11929}, 2020.

\end{thebibliography}
\clearpage
\appendix

% \section{Limitations}
% \xhdr{Markov Assumption}
% 3 fold repetition

% \xhdr{Game Filtering}
% While Maia-2 has been trained exclusively on Lichess Rapid games to ensure a coherent analysis within a consistent rating system, the adaptability of its architecture also opens avenues for other categories of online chess games. Bullet, Blitz and Classical games offer rich datasets for furthering our understanding of human decision-making under various time constraints. To inherit the capabilities of Maia-2 to these domain, dedicated models catering for other rating systems would be required for a new training. However, this separation in game types would best preserve the integrity of the rating-specific treats in our model.

%\subsection{Rapid Only}

\begin{figure*}[!t]
% \vspace{-0.4cm}
	\centering
	\includegraphics[width=0.95\textwidth]{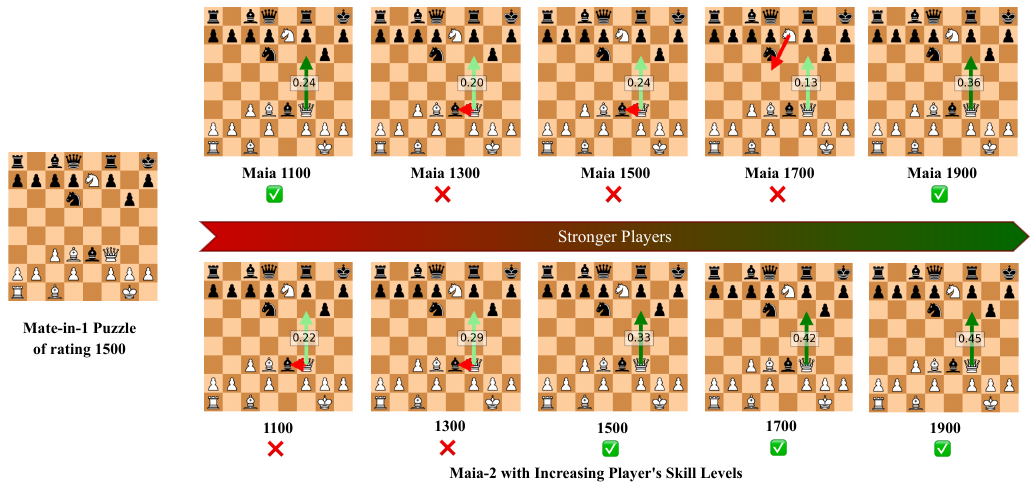}
%  \vspace{-0.4cm}
        \caption{Maia-2 and Maia-1 solving a Mate-in-1 chess puzzle of rating 1500. Green arrows represent correct move predictions, while red arrows indicate incorrect predictions. The darkness of the green color correlates with the model's confidence, with darker arrows denoting a higher probability of making the correct move.}
  \label{fig:teaser}
% \vspace{-0.6cm}
\end{figure*}

\section{More Experimental Results}
\label{appendix:more_results}

\xhdr{Case study: Smoothness} 
We evaluate the smoothness of Maia-1 and Maia-2 by a case study in puzzle solving: a Mate-in-1 puzzle of 1500 skill level is presented to both Maia-1 and Maia-2, where smoothness can be evaluated by checking whether the predictions are \emph{monotonic} and \emph{transitional} as skill level increases. Call a model's treatment of a position as \emph{monotonic} if the predicted probability of the correct move increases with skill monotonically, we can observe from Figure~\ref{fig:teaser} that the Maia-2 predicted probabilities of the best move (in green arrows) increase monotonically from 0.22 to 0.45 as the skill levels rise from 1100 to 1900, while Maia-1 predictions are rather turbulent. 
Similarly, we call a model's treatment of a position \emph{transitional} if it predicts a suboptimal move for some prefix of skills and then transitions to an optimal move for all subsequent skill levels. 
As shown in Figure~\ref{fig:teaser}, Maia-2 can mimic weaker players to whom the puzzle is hard to solve, while stronger Maia-2 with skill level configured above or equal to 1500 can successfully solve the puzzle. However, Maia 1100 surprisingly solved the puzzle, while the stronger Maia-1 models, e.g., Maia 1700 failed to make the optimal move. 
Therefore, in the considered case, as opposed to Maia-1, Maia-2 yields smooth predictions provided that its treatment of this position is \emph{monotonic} and \emph{transitional}.
% The by construction no-lookback attribute of Maia-2 also favors the model in playing chess puzzles, where only one move is provided as the move history, while by nature the dependence on move history of Maia-1 fairly deteriorates its performance on puzzles. As a case study, we pick a straightforward Mate-in-1 puzzle of Lichess rating 1500 in Figure \ref{fig:teaser}. The suggested moves of Maia-1 from Maia 1100 to Maia 1900 turn out to be volatile, not showing the pattern that higher-rated Maia-1 are consistently better at dealing with this Mate-in-1 position. In comparison, although Maia-2 when injected with weaker ratings fail at this puzzle as well, from rating 1500 Maia-2 manages to predict the correct move in all cases. Moreover, the probability for Maia-2 to make the correct move consistently rises up as the input skill ratings increase, which is noteworthy especially for a chess model with no search to make decisions.

\begin{figure*}[!t]
% \vspace{-0.4cm}
	\centering
	\includegraphics[width=0.9\textwidth]{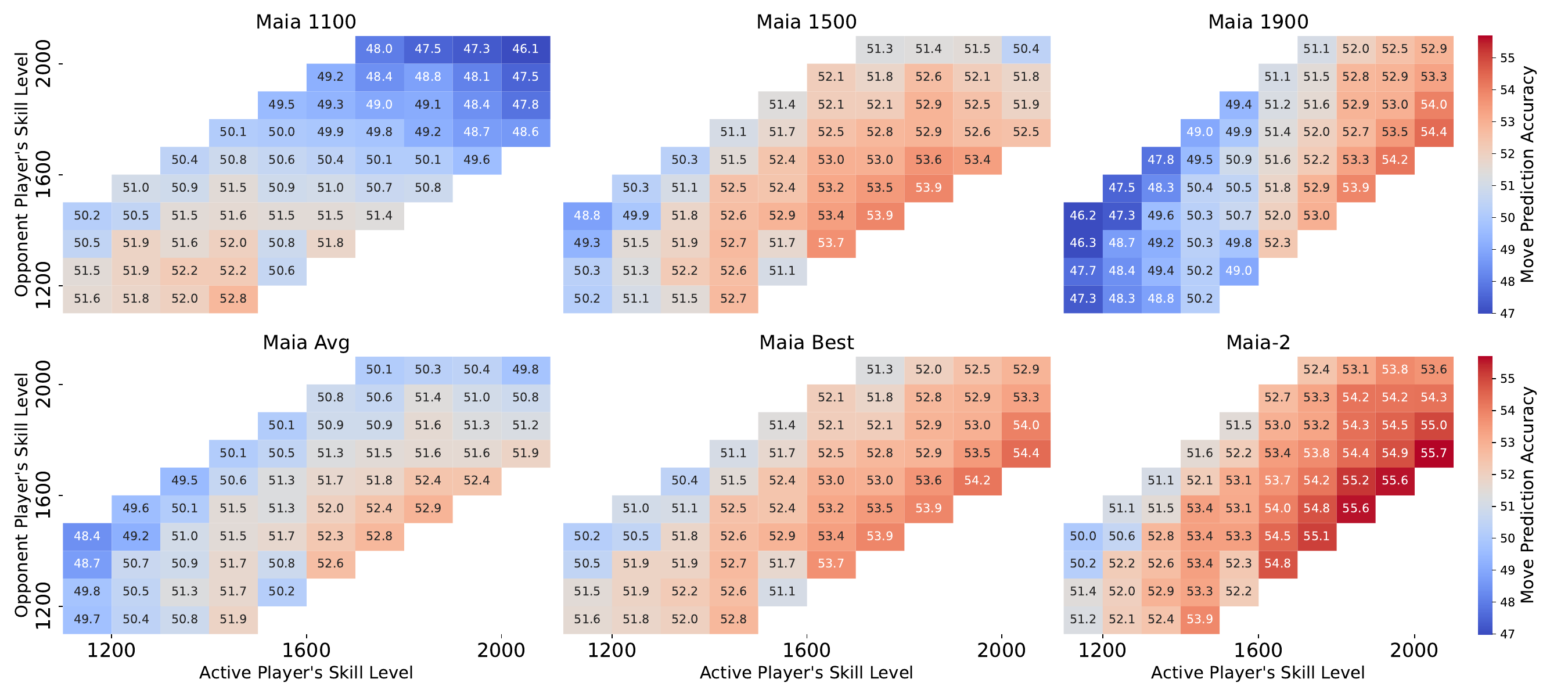}
%  \vspace{-0.4cm}
        \caption{Move prediction accuracy across diverse skill levels. Colors represent performance, with warmer tones indicating higher accuracy. Missing skill combinations were too rare to be statistically reliable. }
  \label{fig:heatmap_appendix}
% \vspace{-0.6cm}
\end{figure*}

\begin{figure}[!t]
% \vspace{-0.4cm}
	\centering
	\includegraphics[width=0.85\textwidth]{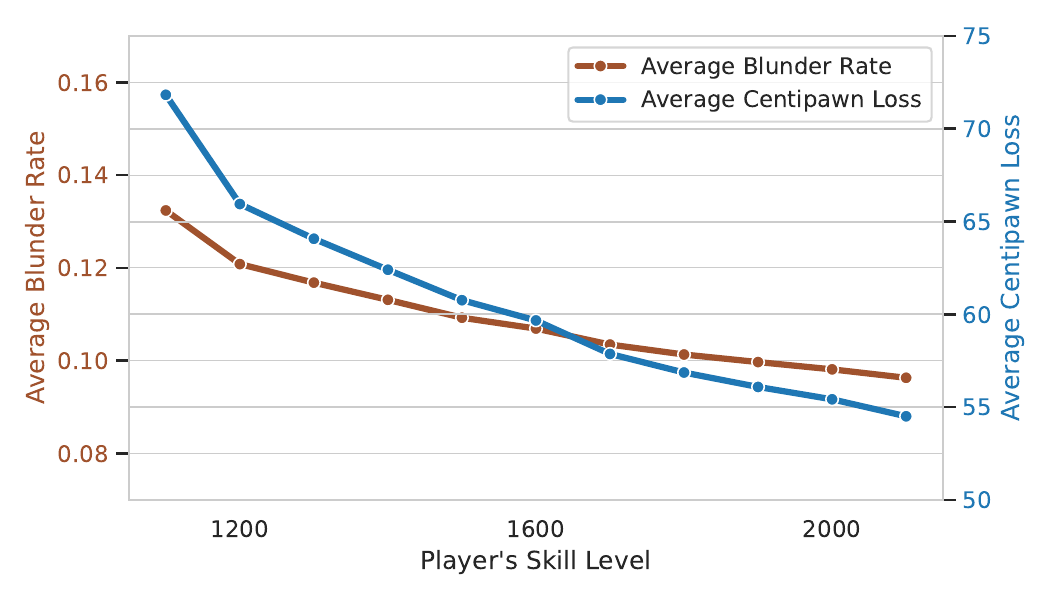}
%  \vspace{-0.4cm}
        \caption{Quality of predicted moves quantified by blunder rate and centipawn Loss.}
  \label{fig:sf_eval}
% \vspace{-0.6cm}
\end{figure}

\xhdr{Quality of predicted moves} Given the same chess position but increasing skill levels, a coherent, skill-aware human move prediction model should make progressively higher-quality moves. We measure the average centipawn loss (a standard move quality metric, the lower the better) and the average blunder rate (fraction of times an egregious mistake is predicted) in positions randomly sampled from December 2023 games. The centipawn loss is determined by comparing the moves predicted by Maia-2 against the top move as evaluated by Stockfish at depth 20 (human grandmaster-level play), while blunders are classified as moves resulting in a win-rate loss of 10\% or more. As shown in Figure~\ref{fig:sf_eval}, both the centipawn loss and the blunder rate exhibit a smooth, monotonic decrease as player skill rises, demonstrating Maia-2's capability of adjusting its predictions coherently to align with the increasingly skilled players.

\begin{figure}[htbp]
% \vspace{-0.4cm}
	\centering
	\includegraphics[width=0.6\textwidth]{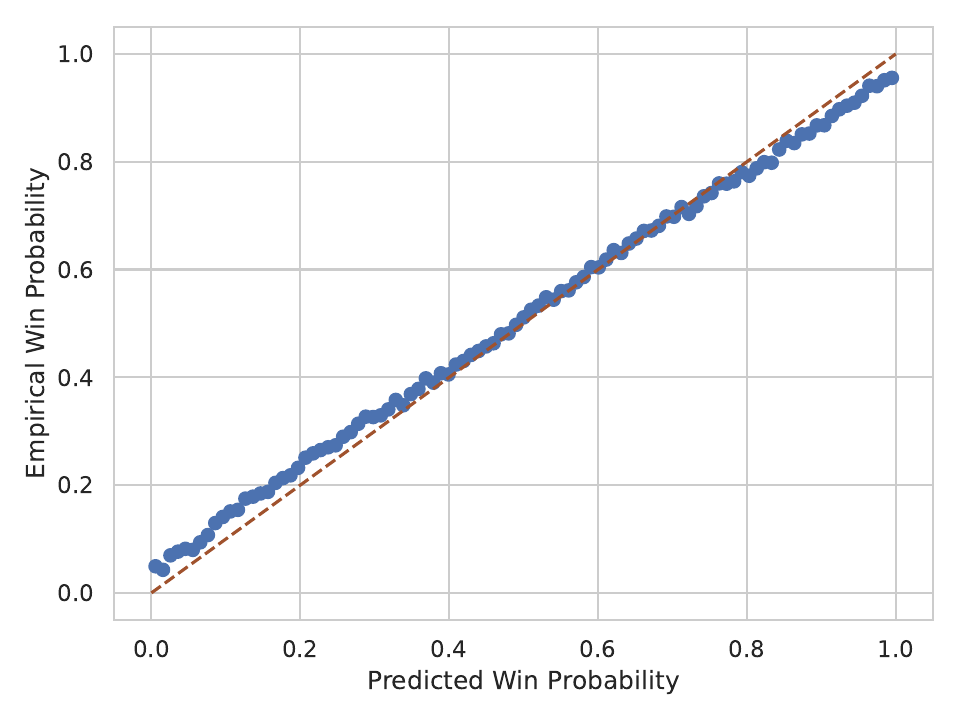}
%  \vspace{-0.4cm}
        \caption{Value head Q-Q plot.}
  \label{fig:qqplot}
% \vspace{-0.6cm}
\end{figure}

\xhdr{Value head} As a proxy of model evaluation given a board position, we train the model value head which is potentially significant for a wide variety of downstream tasks. The value head is trained as a regression task from -1 to 1 indicating from losing to winning positions, and finally normalized to a continuous value between 0 and 1 similar to AlphaZero. The correct label for value head is the actual game results. To check the evaluation quality of our model value head, we calibrate the value head results with actual game outcomes, and generate a quantile-quantile plot in Figure \ref{fig:qqplot} where the win probability is discretized uniformly into 100 bins from 0 to 1.

\xhdr{Ablation study}
To understand which components of our architecture are most responsible for the performance gains, we conduct an ablation study on the \textit{Maia Testset}. We train a version of \modelname{}$_{\text{subset}}$ using a naive method of incorporating encoded skill levels without the proposed skill-aware attention module (``w/o Att''). In this model, the skill level encodings $\mathbf{e}_a$ and $\mathbf{e}o$ are directly concatenated with flattened $P_{\text{encoded}}$, without bridging them with the skill-aware attention (skipping Section~\ref{section:attention}), and directly connected to prediction heads. As shown in Table~\ref{tab:ablation}, \modelname{}$_{\text{subset}}$ consistently performs better with skill-aware attention, demonstrating the necessity of modeling the complexity and non-linearity in player's skill development in sophisticated ways and the effectiveness of our proposed unified modeling approach with skill-aware attention to model such nuances. We also train a version of \modelname{}$_{\text{subset}}$ without the auxiliary information head (``w/o Aux''). These results show that infusing auxiliary information as labels during model training also results in a significant performance improvement, although not as dramatically as skill-aware attention.

\begin{figure}[htbp]
% \vspace{-0.4cm}
	\centering
	\includegraphics[width=0.9\textwidth]{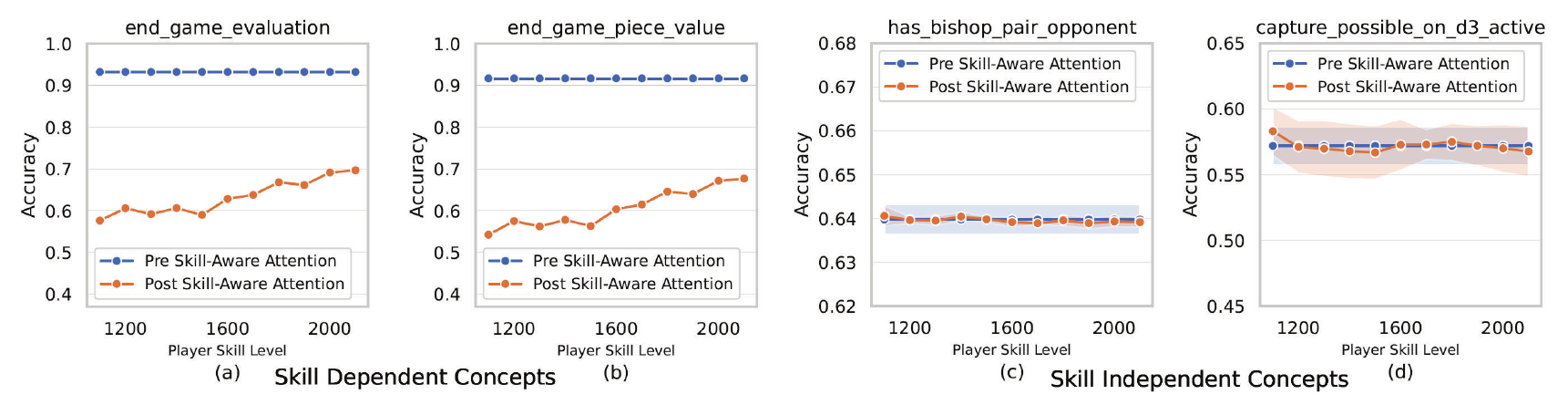}
 % \vspace{-0.4cm}
        \caption{Maia-2's chess concept recognition as a function of skill level, as measured by linear activation probes right before (blue) and after (orange) skill-aware attention. (a) Stockfish overall board evaluation for end-game positions. (b) Stockfish evaluation of end-game bonuses and penalties to pieces for white. (c) Does the opponent player own two bishops? (d) Is capture possible on square d3 for the active player? }
  \label{fig:probing2}
% \vspace{-0.6cm}
\end{figure}

\begin{figure}[htbp]
% \vspace{-0.3cm}
	\centering
	\includegraphics[width=0.7\textwidth]{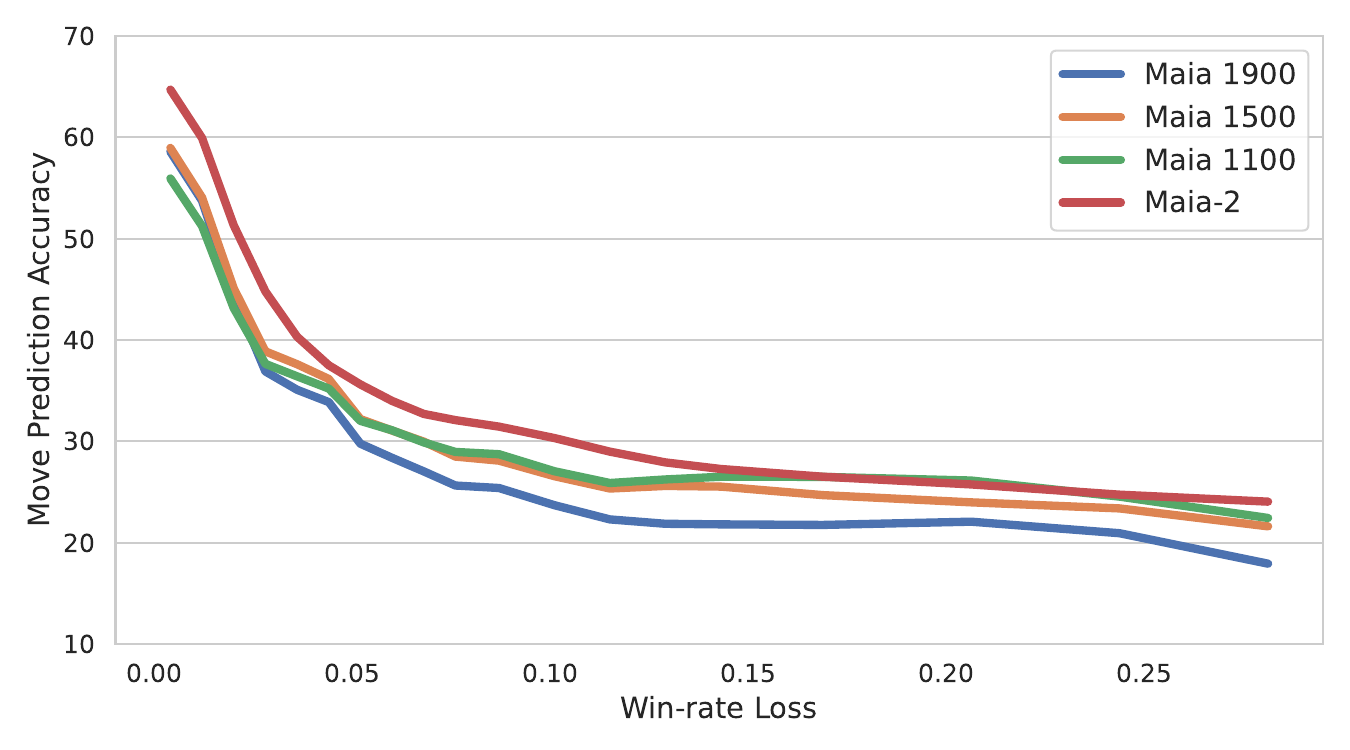}
 % \vspace{-0.4cm}
        \caption{Move prediction accuracy as a function of move quality, quantified by win-rate loss.}
  \label{fig:acc_winprob}
% \vspace{-0.2cm}
\end{figure}

\xhdr{Accuracy as a function of move quality}
Figure~\ref{fig:acc_winprob} shows the move prediction accuracy on the \textit{Grounded Testset} as a function of move quality, measured by win-rate loss, which is calculated following the same procedures as prior studies~\cite{mcilroy2020aligning, mcilroy2022learning}. All models generally decrease in their ability to predict worse moves, since humans are generally trying to avoid mistakes, and high-quality moves are more certain whereas lower-quality moves can be more random and thus hard to predict.
Nevertheless, Maia-2 outperforms all versions of Maia across most of the move quality range, demonstrating the effectiveness of our unified modeling approach for human move prediction. Maia-2's overall gains are not constrained to any specific move quality type, but are spread across the entire range.

\section{Reproductibility}
%\subsection{Hyperparameter Settings}

\begin{table}[htbp]
  \centering
  \caption{Hyperparameter Settings.}
    \begin{tabular}{lll}
    \toprule
    \#Games per chunk $N_{\text{chunk}}$ & 20000 \\
    \#Maximum games per skill level $N_{\text{range}}$ & 20 \\
    Initial learning rate & $1e^{-4}$ \\
    Weight decay & $1e^{-5}$ \\
    Batch size (positions) & 8192 \\
    Minimum move ply & 10 \\
    Maximum move ply &  300  \\
    Remaining seconds threshold & 30 \\
    \#Backbone blocks $K_{Conv}$ & 12 \\
    \#Attention block $K_{Att}$ & 2 \\
    \#Input channels $C_{\text{input}}$ & 18 \\
    \#Intermediate channels $C_{\text{mid}}$ & 256 \\
    \#Encoded channels $C_{\text{patch}}$ & 8 \\
    Skill level embedding dimension $d_s$ & 128 \\
    Attention head dimension $d_h$ & 64 \\
    Attention intermediate dimension $d_{\text{att}}$ & 1024 \\
    \#Attention heads $h$ & 16 \\
    \bottomrule
    \end{tabular}%
  \label{tab:hyperparameter}%
\end{table}%

% Table generated by Excel2LaTeX from sheet 'Statistics'
\begin{table}[htbp]
  \centering
  \caption{Statistics of the \textit{Maia-1 Testset}.}
    \begin{tabular}{ccc}
    \toprule
          & \#Positions & Rating Range \\
\cmidrule{2-3}    Total & 106,740 & - \\
\cmidrule{2-3}
    Skilled & 56,812 & < 1599 \\
    Advanced & 41,747 & 1600 - 1999 \\
    Master & 8,181 & $\geq$ 2000 \\
    \bottomrule
    \end{tabular}%
  \label{tab:maia1data_stats}%
\end{table}%

% Table generated by Excel2LaTeX from sheet 'Statistics'
\begin{table}[htbp]
  \centering
  \caption{Statistics of training datasets.}
    \begin{tabular}{lll}
    \toprule
          & Maia-2$_{\text{subset}}$ & Maia-2 \\
\cmidrule{2-3}    \#Games Consumed & 930.19M & 5.14B \\
    \#Games Trained & 21.31M & 168.93M \\
    \#Positions Trained & 1.18B & 9.15B \\
    \bottomrule
    \end{tabular}%
  \label{tab:trainingdata_stats}%
\end{table}%

\xhdr{Implementation details}
\label{appendix:setup}
\label{appendix:detail}
To maintain a consistent perspective from both sides of players, we implemented board flipping to train and test Maia-2; that is, positions with black to move were mirrored such that all analyses could be conducted from the white side's viewpoint. We further refined our dataset through game and position filtering, selecting only rapid games from Lichess with available clock information and disregarding the initial 10 plys of each game as well as positions where either player had less than thirty seconds remaining. The filtration is significant for eliminating the noise introduced by rushed decisions under time constraints, which could skew the true representation of a player's skill. The choice of exclusive rapid games is also informed by the distinct rating systems across different game types, which necessitates the separation of data to maintain rating consistency. We report all hyperparameters involved in training Maia-2 in Table \ref{tab:hyperparameter}. It took approximately 13 days to train Maia-2 with 2$\times$A100 (80G) GPUs under our default settings.

\xhdr{Data Balancing}
\label{appendix:data_balancing}
Chess games between players of significantly different skill levels are relatively rare but help us understand how players of lower skill levels  approach games against far stronger opponents and vice versa.
While previous work has ignored these games completely, they play a central role in our approach. 
Since games between players of similar skill levels vastly outnumber more uneven matchups, we use a data balancing strategy to effectively train our unified model for aligning players across all skill levels, in which games between players of different skill levels are over-sampled. 
To be precise, we pre-process the data in chunks of $N_{\text{chunk}}$ games each. %Since the pre-processing procedure, including data balancing, is computationally expensive, we split the dataset into smaller chunks with $N_{\text{chunk}}$ games per chunk to facilitate multiprocessing. 
We then scan each data chunk to find games satisfying various (active player skill, opponent skill) combinations. Each skill combination can include at most $N_{\text{range}}$ games. We continue scanning the data chunk until all the skill combinations have $N_{\text{range}}$ games or the data chunk is fully consumed. 
Note that the higher the balancing factor $\frac{N_{\text{range}}}{N_{\text{chunk}}}$, the less likely it is that rare skill combinations will fully reach $N_{\text{range}}$, which will lead to less balanced data overall. On the other hand, if the balancing factor is too small, fewer games will be selected from each data chunk, which is data-inefficient. We choose a fair compromise between data efficiency and balanced data so that our training data encompasses a broad spectrum of skill levels without biasing excessively towards the more frequently-occurring equal-skill matchups.

\xhdr{Data Filtering}
Online chess platforms feature a variety of game types, including blitz, rapid, and classical, each representing games played at different time controls (amount of time given to each player for the whole game). 
We use data from \href{https://lichess.org/}{Lichess}, a well-known large open-source chess platform, and its \href{https://database.lichess.org/}{open database}. In Lichess,
since each game type is given a separate rating, ratings across different game types are not comparable (e.g.\ a rating of 1800 in ``Rapid'' is significantly weaker than a rating of 1800 in ``Blitz'' on Lichess). Previous work~\cite{mcilroy2020aligning, mcilroy2022learning} mixes player ratings across these game types together for training and evaluation. %However, identical ratings across different game types can represent different player skill levels\cite{van2007effects}. 
Instead of mixing data across game types, we focus on Rapid games only, which are medium-length games that lie between the fast-paced decisions of ``Blitz'' games and the slower, more strategic considerations of ``Classical'' games. ``Blitz'' and ``Bullet'' games, characterized by their quick pace, are composed of many decisions made under time pressure, introducing randomness that may not accurately reflect player intentions and skills. In contrast, Classical games are played less frequently, leading to data scarcity. Rapid chess is an ideal compromise between quality of play and quantity of data. In addition, we follow the procedures in~\cite{mcilroy2020aligning} to filter valid positions within each game.

%board flipping\\
%game filtering\\
%position filtering \\
%ResNet-based backbone \\
%rapid only \\
\xhdr{Chess position representation}
\label{appendix:channels}
We follow the well-established prior works~\cite{silver2017mastering, mcilroy2020aligning} to represent chess positions as multi-channel $8\times8$ matrices, including:
\begin{itemize}
    \item Piece Representation: The first 12 channels categorize the board's pieces by type and color, with one channel each for white and black Pawns, Knights, Bishops, Rooks, Queens, and Kings. A cell is marked 1 to denote the presence of a piece in the corresponding location, and 0 otherwise.
    \item Player's Turn: A single channel (the 13th) indicates the current player's turn, filled entirely with 1s for white and 0s for black, providing the model with context on whose move is being evaluated.
    \item Castling Rights: Four channels (14th to 17th) encode the castling rights for both players, with the entire channel set to 1 if the right is available or 0 otherwise.
    \item En Passant Target: The final channel (18th) marks the square available for en passant capture, if any, with 1 and 0s elsewhere.
\end{itemize}

\xhdr{Chess concepts probing}
Given a board position, we vary the skill level injected to Maia-2 to extract the $P_{\text{encoded}} \in \mathbb{R}^{C_{\text{patch}}\times8\times8}$ as the learned respresentation of the ResNet-based backbone served as the control group which remain constant to varying skill ratings, and the output hidden states $ P\in \mathbb{R}^{C_{\text{patch}}\times d_{\text{att}}}$ after skill-aware attention directly connected to model heads. We randomly pick 500,000 positions from Lichess December 2023 database and calculate Stockfish built-in or custom implemented concepts using the Forsyth-Edwards Notation of the positions.

We carefully choose $P_{\text{encoded}}$ and $P$ as internal representations for the reason that $P_{\text{encoded}}$ is the comprehensive understanding of chess board for the backbone network right before skill-aware attention, and $P$ is the final hidden states that directly influence model decision and evaluation. Following the work of Alphazero concept probing ~\cite{mcgrath2022acquisition}, for chess concepts with continuous values we train Lasso regressors with coefficient for $\mathcal{L}^1$-regularization selected by 5-fold cross-validation, and for binary concepts we train logistic regressors with downsampling. During evaluation on unbalanced test set, we measure the performance of continuous-valued probes with the coefficient of determination $r^2$, and score the binary-valued probes with the Macro F1. 

% Table generated by Excel2LaTeX from sheet 'Statistics'
\begin{table*}[htbp]
% \small
\tiny
  \centering
  \caption{Statistics of the \textit{Cross-skill Testset}.}
    \begin{tabular}{cccccccccc}
    \toprule
    \multirow{2}[4]{*}{Opponent} & \multicolumn{9}{c}{Active} \\
\cmidrule{2-10}          & 1100 - 1199 & 1200 - 1299 & 1300 - 1399 & 1400 - 1499 & 1500 - 1599 & 1600 - 1699 & 1700 - 1799 & 1800 - 1899 & $\geq$2000 \\
    \midrule
    $\geq$2000  & -     & -     & -     & -     & -     & 91,264 & 100,000 & 100,000 & 100,000 \\
    1800 - 1899      & -     & -     & -     & -     & 83,337 & 99,931 & 100,000 & 100,000 & 100,000 \\
    1700 - 1799     & -     & -     & -     & 91,054 & 100,000 & 100,000 & 100,000 & 100,000 & 100,000 \\
    1600 - 1699      & -     & -     & 81,386 & 99,998 & 100,000 & 100,000 & 100,000 & 99,933 & 91,610 \\
    1500 - 1599     & -     & 82,059 & 99,500 & 100,000 & 100,000 & 100,000 & 99,974 & 83,650 & - \\
    1400 - 1499     & 89,666 & 100,000 & 100,000 & 100,000 & 100,000 & 100,000 & 91,291 & -     & - \\
    1300 - 1399  & 97,602 & 100,000 & 100,000 & 100,000 & 99,544 & 81,692 & -     & -     & - \\
    1200 - 1299  & 100,000 & 100,000 & 100,000 & 100,000 & 82,230 & -     & -     & -     & - \\
    1100 - 1199  & 100,000 & 100,000 & 97,635 & 89,643 & -     & -     & -     & -     & - \\
    \bottomrule
    \end{tabular}%
  \label{tab:maia2data_stats}%
\end{table*}%

\begin{table*}[htbp]
% \small
\tiny
  	\renewcommand\tabcolsep{2pt}
  \centering
  \caption{Number of games in the balanced dataset used for training Maia-2$_{\text{subset}}$.}
    \begin{tabular}{cccccccccccc}
    \toprule
    \multirow{2}[4]{*}{Black} & \multicolumn{11}{c}{White} \\
\cmidrule{2-12}          & <1100 & 1100 - 1199 & 1200 - 1299 & 1300 - 1399 & 1400 - 1499 & 1500 - 1599 & 1600 - 1699 & 1700 - 1799 & 1800 - 1899 & 1900 - 1999 & >=2000 \\
    \midrule
    <1100 & 583,437 & -     & -     & -     & -     & -     & -     & -     & -     & -     & - \\
    1100 - 1199 & 583,403 & 583,433 & -     & -     & -     & -     & -     & -     & -     & -     & - \\
    1200 - 1299 & 427,160 & 583,402 & 583,443 & -     & -     & -     & -     & -     & -     & -     & - \\
    1300 - 1399 & 176,930 & 408,819 & 583,421 & 583,443 & -     & -     & -     & -     & -     & -     & - \\
    1400 - 1499 & 113,683 & 204,425 & 499,185 & 583,433 & 583,445 & -     & -     & -     & -     & -     & - \\
    1500 - 1599 & 101,756 & 139,741 & 307,446 & 559,722 & 583,440 & 583,439 & -     & -     & -     & -     & - \\
    1600 - 1699 & 57,381 & 85,483 & 174,417 & 384,670 & 566,851 & 583,436 & 583,435 & -     & -     & -     & - \\
    1700 - 1799 & 32,621 & 47,671 & 100,841 & 188,967 & 359,056 & 571,666 & 583,421 & 583,428 & -     & -     & - \\
    1800 - 1899 & 16,656 & 26,195 & 52,043 & 105,095 & 178,957 & 348,979 & 557,921 & 583,415 & 583,422 & -     & - \\
    1900 - 1999 & 8,032 & 11,736 & 23,800 & 48,965 & 84,808 & 150,182 & 272,451 & 529,071 & 583,388 & 582,715 & - \\
    >2000 & 5,927 & 7,953 & 14,825 & 28,929 & 49,237 & 88,317 & 134,432 & 280,452 & 531,162 & 582,993 & 582,542 \\
    \bottomrule
    \end{tabular}%
  \label{tab:balanceddata_stats_subset}%
\end{table*}%

% Table generated by Excel2LaTeX from sheet 'Statistics'
\begin{table*}[htbp]
\tiny
  	\renewcommand\tabcolsep{2pt}
  \centering
  \caption{Number of games in the balanced dataset used for training Maia-2.}
    \begin{tabular}{cccccccccccc}
    \toprule
    \multirow{2}[4]{*}{Black} & \multicolumn{11}{c}{White} \\
\cmidrule{2-12}          & <1100 & 1100 - 1199 & 1200 - 1299 & 1300 - 1399 & 1400 - 1499 & 1500 - 1599 & 1600 - 1699 & 1700 - 1799 & 1800 - 1899 & 1900 - 1999 & >=2000 \\
    \midrule
    <1100 & 4,698,087 & -     & -     & -     & -     & -     & -     & -     & -     & -     & - \\
    1100 - 1199 & 4,697,937 & 4,697,979 & -     & -     & -     & -     & -     & -     & -     & -     & - \\
    1200 - 1299 & 3,165,114 & 4,697,972 & 4,698,076 & -     & -     & -     & -     & -     & -     & -     & - \\
    1300 - 1399 & 2,015,110 & 2,561,344 & 4,698,015 & 4,698,073 & -     & -     & -     & -     & -     & -     & - \\
    1400 - 1499 & 1,509,456 & 1,549,218 & 3,008,520 & 4,698,033 & 4,698,093 & -     & -     & -     & -     & -     & - \\
    1500 - 1599 & 1,259,824 & 1,402,474 & 2,348,863 & 3,895,270 & 4,698,047 & 4,698,106 & -     & -     & -     & -     & - \\
    1600 - 1699 & 627,583 & 780,364 & 1,415,147 & 2,257,063 & 3,514,829 & 4,698,065 & 4,698,092 & -     & -     & -     & - \\
    1700 - 1799 & 419,889 & 408,444 & 891,818 & 1,499,773 & 2,228,952 & 3,999,818 & 4,698,035 & 4,698,088 & -     & -     & - \\
    1800 - 1899 & 281,902 & 286,769 & 447,341 & 893,978 & 1,468,987 & 2,439,832 & 3,665,572 & 4,698,000 & 4,698,059 & -     & - \\
    1900 - 1999 & 190,385 & 171,133 & 272,258 & 448,351 & 818,918 & 1,452,505 & 1,940,287 & 3,486,275 & 4,697,958 & 4,697,311 & - \\
    >2000 & 228,477 & 186,001 & 277,431 & 423,692 & 651,232 & 1,206,468 & 1,718,683 & 2,587,573 & 3,967,891 & 4,697,533 & 4,697,178 \\
    \bottomrule
    \end{tabular}%
  \label{tab:balanceddata_stats}%
\end{table*}%

\clearpage
\section*{NeurIPS Paper Checklist}

%%% BEGIN INSTRUCTIONS %%%
The checklist is designed to encourage best practices for responsible machine learning research, addressing issues of reproducibility, transparency, research ethics, and societal impact. Do not remove the checklist: {\bf The papers not including the checklist will be desk rejected.} The checklist should follow the references and follow the (optional) supplemental material.  The checklist does NOT count towards the page
limit. 

Please read the checklist guidelines carefully for information on how to answer these questions. For each question in the checklist:
\begin{itemize}
    \item You should answer \answerYes{}, \answerNo{}, or \answerNA{}.
    \item \answerNA{} means either that the question is Not Applicable for that particular paper or the relevant information is Not Available.
    \item Please provide a short (1–2 sentence) justification right after your answer (even for NA). 
   % \item {\bf The papers not including the checklist will be desk rejected.}
\end{itemize}

{\bf The checklist answers are an integral part of your paper submission.} They are visible to the reviewers, area chairs, senior area chairs, and ethics reviewers. You will be asked to also include it (after eventual revisions) with the final version of your paper, and its final version will be published with the paper.

The reviewers of your paper will be asked to use the checklist as one of the factors in their evaluation. While "\answerYes{}" is generally preferable to "\answerNo{}", it is perfectly acceptable to answer "\answerNo{}" provided a proper justification is given (e.g., "error bars are not reported because it would be too computationally expensive" or "we were unable to find the license for the dataset we used"). In general, answering "\answerNo{}" or "\answerNA{}" is not grounds for rejection. While the questions are phrased in a binary way, we acknowledge that the true answer is often more nuanced, so please just use your best judgment and write a justification to elaborate. All supporting evidence can appear either in the main paper or the supplemental material, provided in appendix. If you answer \answerYes{} to a question, in the justification please point to the section(s) where related material for the question can be found.

IMPORTANT, please:
\begin{itemize}
    \item {\bf Delete this instruction block, but keep the section heading ``NeurIPS paper checklist"},
    \item  {\bf Keep the checklist subsection headings, questions/answers and guidelines below.}
    \item {\bf Do not modify the questions and only use the provided macros for your answers}.
\end{itemize}

%%% END INSTRUCTIONS %%%

\begin{enumerate}

\item {\bf Claims}
    \item[] Question: Do the main claims made in the abstract and introduction accurately reflect the paper's contributions and scope?
    \item[] Answer: \answerYes{} % Replace by , \answerNo{}, or \answerNA{}.
    \item[] Justification: The abstract and introduction clearly outline the main contributions and scope of the paper.

\item {\bf Limitations}
    \item[] Question: Does the paper discuss the limitations of the work performed by the authors?
    \item[] Answer: \answerYes{} % Replace by \answerYes{}, \answerNo{}, or \answerNA{}.
    \item[] Justification: We discuss the limitations of our work in Section 5.

\item {\bf Theory Assumptions and Proofs}
    \item[] Question: For each theoretical result, does the paper provide the full set of assumptions and a complete (and correct) proof?
    \item[] Answer: \answerNA{} % Replace by \answerYes{}, \answerNo{}, or \answerNA{}.
    \item[] Justification: We do not include theoretical results.

    \item {\bf Experimental Result Reproducibility}
    \item[] Question: Does the paper fully disclose all the information needed to reproduce the main experimental results of the paper to the extent that it affects the main claims and/or conclusions of the paper (regardless of whether the code and data are provided or not)?
    \item[] Answer: \answerYes{} % Replace by \answerYes{}, \answerNo{}, or \answerNA{}.
    \item[] Justification: We include details of model design and model training in Section 3, experimental setup in Section 4, as well as hyperparameter settings and implementation details in Appendix B.

\item {\bf Open access to data and code}
    \item[] Question: Does the paper provide open access to the data and code, with sufficient instructions to faithfully reproduce the main experimental results, as described in supplemental material?
    \item[] Answer: \answerYes{} % Replace by \answerYes{}, \answerNo{}, or \answerNA{}.
    \item[] Justification: We include the link to our code including the data processing pipeline in the abstract.

\item {\bf Experimental Setting/Details}
    \item[] Question: Does the paper specify all the training and test details (e.g., data splits, hyperparameters, how they were chosen, type of optimizer, etc.) necessary to understand the results?
    \item[] Answer: \answerYes{} % Replace by \answerYes{}, \answerNo{}, or \answerNA{}.
    \item[] Justification: We include dataset details and hyperparameter settings in Section 4 and Appendix B.

\item {\bf Experiment Statistical Significance}
    \item[] Question: Does the paper report error bars suitably and correctly defined or other appropriate information about the statistical significance of the experiments?
    \item[] Answer: \answerNo{} % Replace by \answerYes{}, \answerNo{}, or \answerNA{}.
    \item[] Justification: We train Maia-2 with a huge amount (9.1B) of chess positions. Therefore, it is hard to evaluate Maia-2 multiple times with different train/test splits.

\item {\bf Experiments Compute Resources}
    \item[] Question: For each experiment, does the paper provide sufficient information on the computer resources (type of compute workers, memory, time of execution) needed to reproduce the experiments?
    \item[] Answer: \answerYes{} % Replace by \answerYes{}, \answerNo{}, or \answerNA{}.
    \item[] Justification: We report the required computational resources in Appendix B.
    
\item {\bf Code Of Ethics}
    \item[] Question: Does the research conducted in the paper conform, in every respect, with the NeurIPS Code of Ethics \url{https://neurips.cc/public/EthicsGuidelines}?
    \item[] Answer: \answerYes{} % Replace by \answerYes{}, \answerNo{}, or \answerNA{}.
    \item[] Justification: We conduct our work under the guidance of NeurIPS Code of Ethics.

\item {\bf Broader Impacts}
    \item[] Question: Does the paper discuss both potential positive societal impacts and negative societal impacts of the work performed?
    \item[] Answer: \answerYes{} % Replace by \answerYes{}, \answerNo{}, or \answerNA{}.
    \item[] Justification: We discuss the potential social impacts of our work in Section 1 and Section 5.
    
\item {\bf Safeguards}
    \item[] Question: Does the paper describe safeguards that have been put in place for responsible release of data or models that have a high risk for misuse (e.g., pretrained language models, image generators, or scraped datasets)?
    \item[] Answer: \answerYes{} % Replace by \answerYes{}, \answerNo{}, or \answerNA{}.
    \item[] Justification: We discuss the limited risks of our work as a human-like model in Section 5.

\item {\bf Licenses for existing assets}
    \item[] Question: Are the creators or original owners of assets (e.g., code, data, models), used in the paper, properly credited and are the license and terms of use explicitly mentioned and properly respected?
    \item[] Answer: \answerYes{} % Replace by \answerYes{}, \answerNo{}, or \answerNA{}.
    \item[] Justification: We include all assets used in our paper in main texts and references.

\item {\bf New Assets}
    \item[] Question: Are new assets introduced in the paper well documented and is the documentation provided alongside the assets?
    \item[] Answer: \answerYes{} % Replace by \answerYes{}, \answerNo{}, or \answerNA{}.
    \item[] Justification: We release our code with documentations and comments.

\item {\bf Crowdsourcing and Research with Human Subjects}
    \item[] Question: For crowdsourcing experiments and research with human subjects, does the paper include the full text of instructions given to participants and screenshots, if applicable, as well as details about compensation (if any)? 
    \item[] Answer: \answerNA{} % Replace by \answerYes{}, \answerNo{}, or \answerNA{}.
    \item[] Justification: Our work does not involve crowdsourcing.

\item {\bf Institutional Review Board (IRB) Approvals or Equivalent for Research with Human Subjects}
    \item[] Question: Does the paper describe potential risks incurred by study participants, whether such risks were disclosed to the subjects, and whether Institutional Review Board (IRB) approvals (or an equivalent approval/review based on the requirements of your country or institution) were obtained?
    \item[] Answer: \answerNA{} % Replace by \answerYes{}, \answerNo{}, or \answerNA{}.
    \item[] Justification: Our work does not involve crowdsourcing.

\end{enumerate}

\end{document}